\documentclass[twocolumn]{svjour3}
\smartqed
\usepackage{graphicx}
\usepackage{bm}
\usepackage{amssymb}
\usepackage{amsmath}
\usepackage{color}
\newcommand{\subfiglabelright}[1]{%
   \hbox to0pt{\small\kern-2em\hfill\raisebox{.5em}{%
   \colorbox{white}{\hbox to0.7em{\rule{0pt}{0.7em}%
   \hfill\smash{\textbf{#1}}\hfill}}\kern.2em}}}
\begin{document}
\title{Fast and Robust Linear Motion Deblurring}
\author{Martin Welk \and 
        Patrik Raudaschl \and 
        Thomas Schwarzbauer \and
        Martin Erler \and
        Martin L\"auter}
\authorrunning{M. Welk, P. Raudaschl, T. Schwarzbauer,
M. Erler, M. L\"auter}
\institute{%
M. Welk, P. Raudaschl, T. Schwarzbauer, M. Erler \at
Institute for Biomedical Image Analysis,
University for Medical Informatics and Technology (UMIT), 
6060 Hall/Tyrol, Austria\\ 
Tel.: +43-50-8648-3974\\
\email{martin.welk@umit.at, patrik.raudaschl@umit.at, 
thomas.schwarzbauer@umit.at, martin.erler@umit.at}
\and
M. L\"auter \at
TOMAXX GmbH, 85609 Aschheim, Germany\\
\email{laeuter@googlemail.com}
}
\date{}
\maketitle

\begin{abstract}
We investigate efficient algorithmic realisations for robust 
deconvolution of grey-value images with known space-invariant 
point-spread function, with emphasis on 1D motion blur scenarios. 
The goal is to make deconvolution suitable as 
preprocessing step in automated image processing environments
with tight time constraints.
Candidate deconvolution methods are selected for their restoration
quality, robustness and efficiency. Evaluation of restoration
quality and robustness on synthetic and real-world test images
leads us to focus on a combination of Wiener filtering with few 
iterations of robust and regularised Richardson-Lucy deconvolution.
We discuss algorithmic optimisations for specific scenarios. 
In the case of uniform
linear motion blur in coordinate direction, it is possible to
achieve real-time performance (less than 50 ms) in single-threaded
CPU computation on images of $256\times256$ pixels. 
For more general space-invariant blur settings, still favourable 
computation times are obtained.
Exemplary parallel implementations demonstrate that the proposed
method also achieves real-time performance for general 1D motion blurs 
in a multi-threaded CPU setting, and for general 2D blurs on a GPU.
\end{abstract}
\section{Introduction}
Besides noise, blur is perhaps the most common type of degradation 
in a wide range of imaging modalities. Sources of blur vary widely,
from atmospheric perturbations via optical aberrations, motion of 
both objects and imaging appliances down to defocussing.
Common to all kinds of blur is that it interferes with the 
detection of localised image features, and thereby impedes 
interpretation of images, be it by humans or by automated systems.

It has therefore been a long-standing goal of image processing 
research to remove blur from so degraded images. For this task, 
\emph{deconvolution,} numerous approaches have been designed that 
differ in the requirements they impose on the input data, the 
quality of results, and the computational effort. First work in 
the context of 1D signal processing goes back almost eighty years 
\cite{vanCittert-ZPhys33}. Approaches studied since then range 
from fast linear filters and Fourier-based methods like the Wiener
filter \cite{Wiener-Book49} to non-linear iterative
schemes derived from variational and/or statistical models
\cite{Bar-scs05,Dey-MRT06,%
Krishnan-nips09,%
Lucy-AJ74,Richardson-JOSA72,%
Wang-SIIMS08,%
Welk-tr10}.
All sorts of deconvolution problems are highly ill-posed, and
there is inevitably a trade-off between restoration quality and 
speed of computation.

\paragraph{Basic classification of deconvolution tasks.}
In the classification of deconvolution problems, the most important
dichotomy is that between non-blind and blind deconvolution.
Whereas in the first case the blur is known, blind deconvolution
assumes that only the observed image is available, and the blur is 
to be estimated along with the sharp image. 
As blind deconvolution is much more underconstrained than 
deconvolution with known blur, it involves higher 
computational cost and often generates poorer results.

A second distinction is that between space-invariant blur, in which 
the so-called \emph{point-spread function} is the same for all 
locations 
in the image domain, and space-variant blur, where different pixels
are smeared in different ways. In the space-invariant case, the 
blur operator reduces to a standard convolution, such that Fourier 
methods can be used, and also forward (blur) operations in iterative
schemes can be computed by the Fast Fourier Transform.

\paragraph{Towards real-time deconvolution.}
In this contribution, we consider robust deconvolution 
with known space-invariant blur under tight time constraints, 
with the aim to achieve real-time 
performance at least in some cases. We will therefore
select deconvolution methods for robust deconvolution
performance, and discuss measures to optimise their algorithmic
efficiency.

Real-time deconvolution has also been considered in some
recent papers, see
\cite{Hirsch-iccv11,Klosowski-spie11}. Both papers address
the case of general 2D blur on the basis of the algorithm from
\cite{Krishnan-nips09}, and reach real-time performance when
using GPUs with several hundred computing units under CUDA.
In \cite{Hirsch-iccv11}, even blind deconvolution is considered;
however, the blind deconvolution case is far from real-time
performance even when computing on a GPU.

Comparing to these contributions, however, our goal is different.
We want to investigate under which conditions real-time 
performance can be achieved in \emph{single-threaded CPU} 
computation. This was an essential requirement of the
application context from which this work emerged.
Additional work on multi-threaded CPU and 
GPU realisations is presented to put the results in context.

To accomplish this task, we are willing, in turn, to sacrifice
some generality: We focus on setups in which some parameters can
be chosen in a way such as to reduce the computational 
expense. For example, we accept the limitation to power-of-two
image dimensions such as to profit best from Fast Fourier Transform.
Also, we consider settings restricted to either one-di\-men\-sio\-nal 
blur, or even more specifically just linear motion blur, and assume 
herein that the camera system can be adjusted in such a way that the
1D blur is in the direction of a coordinate axis of the sensor.

\paragraph{Robustness.}
On the other hand, with practical applications of deconvolution as
goal, robustness is a crucial aspect for our investigation.
The concept of robustness originates from statistics \cite{Huber-book81}
and refers to estimators that are as insensitive as possible to outliers
and model violations. In the case of deconvolution,
this includes not only strong noise (e.g., salt-and-pepper noise in
\cite{Bar-scs05}) but also imprecise blur estimates, or errors
near the image boundary caused by blurring across the boundary, see
\cite{Welk-dagm05}.

There is a long tradition of testing deconvolution algorithms
with synthetically blurred images. Many papers, including
\cite{Klosowski-spie11,Krishnan-nips09,You-icip96}, use only
test cases of this kind. Often also the noise levels being
considered are fairly low (in the range of the quantisation 
noise already caused by discretising grey-values to integers in 
$[0,255]$). 
Such test scenarios have their merit
because they allow to measure reconstruction error against 
a ground truth. However, it is particularly dangerous in the
case of such a severely ill-posed problem like deconvolution
to rely \emph{only} on tests with synthetic blur. Methods that
perform well on such data do often not live up to their promises
when applied to real-world images whose blur or noise deviates
slightly from the perfect theoretical model underlying the
deconvolution approach. 

\paragraph{Application context.}
The motivation for the work presented in this paper is to 
devise a deconvolution framework that can be used as preprocessing 
step for further automated image processing tasks under real-time 
conditions within industrial processes.
It is important to note that the embedding into an industrial
process also limits the computational resources that are available
for computation. While parallelisation on multicore CPUs and GPUs 
plays an ever-increasing role in current high-performance computing,
an essential constraint of the application context that motivated 
the present work was that not more than one CPU kernel could be
assigned to the image processing task.

The setting under consideration involved imaging moving objects 
by stationary cameras. The resulting motion blur is to be 
compensated by deconvolution in order to allow the detection and 
localisation of features on moving objects as well as recognition 
of shapes. To this end, a sufficient sharpening of edges and lines 
would be necessary. Ringing artifacts should be suppressed 
sufficiently in order to not create false detections. Finally, in 
order to be feasible together with subsequent processing tasks in 
real-time, deconvolution should not take more than about 50 ms of
single-threaded computation on a contemporary standard CPU.

In addressing this problem, setups of different degree of 
generality were considered: firstly, blur by uniform linear motion;
secondly, blur by non-uniform linear motion; and 
thirdly, a general 2D blur model. In the linear motion 
cases, it was assumed that the blur direction would be aligned 
with a coordinate axis of the imaging system by technical measures.
In all cases, it was assumed that the imaging conditions would be 
sufficiently controllable to ensure that the blur would be known, 
so no blind deconvolution was considered.

\section{Mathematical Models for Deconvolution}
To describe deconvolution models that were investigated in order
to accomplish the task outlined above,
we start from the common spatially invariant blur model for 
grey-value images,
\begin{equation}
f(\bm{x}) = (g * h)(\bm{x}) + n(\bm{x}) \;,
\end{equation}
where $f$ is the observed blurred image, $g$ is the unobservable
sharp image, $h$ the (known) space-invariant point-spread function (PSF),
and $n$ additive noise. By $\bm{x}=(x,y)\in\mathbb{R}^2$ we denote 
the location in the image plane.

In the following, $u$ denotes the processed image that is to 
approximate $g$. In iterative methods, we denote the $k$-th 
iterate by $u^k$. Some methods involve convolution with $h^*$, the 
adjoint of $h$. In our space-invariant setting, $h^*$ is just $h$ 
reflected at the origin, i.e.\ $h^*(\bm{x})=h(-\bm{x})$.

\subsection{Wiener Filter}
The Wiener filter \cite{Wiener-Book49} filters the Fourier 
transform $\hat{f}$ of $f$ to approximately invert the 
multiplication with $\hat{h}$ that corresponds to the convolution 
with $h$; it reads
\begin{equation}
\hat{u} = \frac{\hat{f}\cdot\Bar{\hat{h}}}{\lvert\hat{h}\rvert^2+K}
\end{equation}
with the filter parameter $K>0$ that dampens the amplification of
those frequencies for which $\lvert\hat{h}\rvert$ is zero or small.
Note that $\Bar{\hat{h}}$, the complex conjugate of 
$\hat{h}$, is the Fourier transform of $h^*$.
Generally, the sharpening effect of the filter is the more 
pronounced, the smaller $K$ is chosen. However, smaller $K$ also 
implies stronger amplification of noise, such that larger $K$ must 
be used when strong noise is present. Using a Gaussian noise model,
$K$ can be chosen dependent on the standard deviation of noise such
as to minimise the mean square error of the approximation of $g$ by
$u$. As a single-step method, the Wiener filter is fast, but its 
applicability is limited since it can hardly cope with more severe 
noise, and like all linear methods tends to produce oscillatory
over- and undershoots near contrasted structures, known as ringing 
artifacts.

\subsection{Richardson-Lucy Deconvolution}
One of the most time-proven and popular iterative deconvolution 
method is
Richardson-Lucy (RL) deconvolution \cite{Lucy-AJ74,Richardson-JOSA72}
\begin{equation}
u^{k+1} = \frac{f}{u^k*h}\cdot u^k \;.
\end{equation}
The single parameter of this method is the number of iterations. 
Unlike the Wiener filter, RL requires and preserves positivity of 
grey-values, and is related to a Poisson noise model.
The method features a \emph{semi-convergence} behaviour: 
With increasing number of iterations, sharpness is improved, but
at the same time noise is amplified and leads to divergence after 
an initial convergence phase. 

\subsection{Variational Models for Deconvolution}
A larger class of iterative methods arises from \emph{variational
models.} In these, one aims at minimising energy functionals 
\cite{Bar-scs05,You-icip96}
that combine a so-called data term which penalises deviations
from the blur model with a regulariser enforcing some smoothness
constraint on the image. 
Such an energy functional can read as \cite{Bar-scs05,Welk-dagm05}
\begin{equation}
E[u] = \frac12\int\limits_{\mathbb{R}^2} \varPhi\bigl((f-u*h)^2\bigr) 
+ \alpha\,
\varPsi\bigl(\lvert\bm{\nabla}u\rvert^2\bigr) \mathrm{d}\bm{x}
\label{VariationalE}
\end{equation}
where $\varPhi,\varPsi:\mathbb{R}^+_0\to\mathbb{R}^+$ are 
non-decreasing differentiable functions.
Both contributions are balanced by
a positive weight parameter $\alpha$.

Besides the identity function $\Phi(s^2)=s^2$, which corresponds
to a quadratic error measure, a common choice for the penaliser 
function $\Phi$ is the regularised $L^1$ error 
$\Phi(s^2)=\sqrt{s^2+\varepsilon^2}$ with a small $\varepsilon>0$, 
see e.g.\ \cite{Bar-scs05}. 
The same type of function can be used 
for $\Psi$. 
The resulting total variation penaliser \cite{Rudin-PHYSD92} 
is popular in image processing due to its favourable edge-preserving
behaviour. In order to even enhance edges, one can use even 
penalisers $\Psi(s^2)$ that are non-convex w.r.t. $s$, see
\cite{Almeida-TIP10,Welk-TBW-scs05}. Quadratic penalisation in
the regulariser is rarely used nowadays in deblurring because of
its blurring effect that directly counteracts the data term.

The objective function in \cite{Krishnan-nips09}, motivated
there from statistical considerations, is a discrete version of 
the energy functional 
\begin{equation}
E[u] = \int\limits_{\mathbb{R}^2} \frac{\lambda}{2}(f-u*h)^2 
+ \lvert\bm{\nabla}u\rvert^{\alpha} \mathrm{d}\bm{x}
\end{equation}
which obviously amounts to an instance of \eqref{VariationalE}
with the quadratic data penalty $\varPhi(s^2)=s^2$, and 
a non-convex power function as regulariser.
With regard to the specific optimisation that is employed
for minimisation in \cite{Krishnan-nips09}, special values like 
$\alpha=2/3,3/4,4/5$ are preferred.
Also in \cite{Wang-SIIMS08} an energy of this type is minimised
but with total variation regularision, i.e.\ $\alpha=1$.

In \cite{Welk-tr10}, the energy functional
\begin{equation}
E[u] = 
\int\limits_{\mathbb{R}^2} \varPhi\bigl(r_f(u*h)\bigr) 
+ \alpha\,
\varPsi\bigl(\lvert\bm{\nabla}u\rvert^2\bigr) \mathrm{d}\bm{x} \;.
\label{ERRRL}
\end{equation}
is introduced whose data term does not depend on
$(f-u*h)$ but instead of the so-called information divergence
\begin{equation}
\label{infdiv}
r_f(s):=s-f-f\ln(s/f)\;.
\end{equation}
A simpler version of the functional \eqref{ERRRL}
can also be linked to the Richardson-Lucy method \cite{Snyder-TIP92}.

\paragraph{Techniques for energy minimisation.}
A gradient descent for \eqref{VariationalE} can be derived using
the Euler-Lagrange framework. Discretising in time by an Euler
forward scheme with time step size $\tau>0$ one arrives at
the iteration
\begin{align}
u^{k+1}=
u^k+\tau\Bigl(&\left(
\varPhi'\bigl((f-u*h)^2\bigr)\cdot(f-u*h)\right)*h^*
\notag\\*
&+\alpha\,\mathrm{div}\left(\varPsi'\bigl(
\lvert\bm{\nabla}u\rvert^2\bigr)
\cdot\bm{\nabla}u\right)\Bigr)
\label{VariationalGD}
\end{align}
whose update term combines a sharpening term (first summand) that 
acts to reduce the data term with a nonlinear diffusion term 
(second summand) with the diffusivity $\varPsi'$ that enforces image 
regularity, compare \cite{Welk-dagm05}. 
The functions $\varPhi',\varPsi':\mathbb{R}^+_0\to\mathbb{R}^+$ are 
nonincreasing.

Unfortunately, gradient descent is a fairly slow procedure, and thus
presently not a candidate for deconvolution under real-time
or near-real-time conditions.

The half-quadratic optimisation technique used in
\cite{Krishnan-nips09,Wang-SIIMS08} is another approach to 
minimising an energy approach. The nonlinearity in the
regulariser is decoupled from the minimisation of the data
term by means of additional variables and a coupling term,
weighted by an additional parameter $\beta$. Following a 
continuation strategy, $\beta$ is successively increased
during the computation.
For details see
\cite{Krishnan-nips09,Wang-SIIMS08}.

For the energy functional \eqref{ERRRL} a 
po\-si\-ti\-vi\-ty-pre\-ser\-ving
iterative minimisation scheme in the style of the RL method can
be derived \cite{Elhayek-dagm11,Welk-tr10}, which is 
called robust and regularised RL deconvolution (RRRL)
and reads as
\begin{equation}
u^{k+1} = \frac
{\left(W^k\cdot\frac{f}{u^k*h}\right)*h^*
+\alpha\,\left[ D_{\varPsi'}(u^k) \right]_+}
{W^k*h^*
-\alpha\,\left[ D_{\varPsi'}(u^k) \right]_-}
\end{equation}
where
\begin{align}
W^k &:= \varPhi'(r_f(u^k*h))\;,\\
D_{\varPsi'}(u^k) &:= 
\mathrm{div}\left(\varPsi'(\lvert\bm{\nabla}u^k\rvert^2)
\bm{\nabla}u^k\right)\;,\\
[z]_{\pm}&:=\frac12(z\pm\lvert z\rvert) \;.
\end{align}

The numerical realisation of these methods will be considered in
Section~\ref{sec-numimp}.

\medskip
In theory, the different noise models
underlying deconvolution models, such as Gaussian noise for the
Wie\-ner filter, Poisson noise for RL etc., render these models
suitable for distinct application areas. In practical application,
however, often not all sources of noise are known and controllable,
such that it is unavoidable to apply deconvolution methods also to
data that do not perfectly match their respective noise models. 
Indeed, robustness, as mentioned in the introduction, is all about 
models being capable of handling this kind of mismatch.

\section{Evaluation of Deconvolution Quality}
\label{sec-qual}
To assess the suitability of the before-mentioned approaches
for our deconvolution task, we start by evaluating their
restoration quality. Gradient descent is not considered
further because of its slowness mentioned earlier.
Apart from that, efficiency optimisation is still not
the goal at this point. All tests in this section are
therefore based on deconvolution implementations
for general 2D blurs with Fourier convolutions.

\begin{table*}[b]
\caption{\label{tab-snr}%
Signal-to-noise ratios (dB) for the blurred 
from Figs.~\ref{fig-cam4} and \ref{fig-cam4g5-6-4i} and
their deblurred versions, including some methods and
parameter settings not shown in the Figures. 
Parameters $K$ for Wiener filter given in brackets apply also
to WR\textsuperscript3L.
Parameters $\lambda$ for WYYZ given in brackets apply also to KF.
All remaining parameters were fixed to the values mentioned in
the captions of Figs.~\ref{fig-cam4} and \ref{fig-cam4g5-6-4i}.
Further details are described in the text.}
\centerline{%
\begin{tabular}{|@{~}c@{~}||@{~}c@{~}|@{~}c@{~}|@{~}c@{~}|@{~}c@{~}|@{~}c@{~}|@{~}c@{~}|@{~}c@{~}|@{~}c@{~}|@{~}c@{~}|}
\hline
\parbox[t]{0.12\textwidth}{\centering Test}&
\parbox[t]{0.078\textwidth}{\centering Blur-}&
\parbox[t]{0.078\textwidth}{\centering Wiener}&
\parbox[t]{0.078\textwidth}{\centering RL}&
\parbox[t]{0.078\textwidth}{\centering WYYZ}&
\parbox[t]{0.078\textwidth}{\centering KF}&
\multicolumn{3}{c@{~}|@{~}}{RRRL}&
\parbox[t]{0.078\textwidth}{\centering WR\textsuperscript3L}\\
\parbox[t]{0.078\textwidth}{\centering image}&
\parbox[t]{0.078\textwidth}{\centering red}&
\parbox[t]{0.078\textwidth}{\centering filter ($K$)}&
\parbox[t]{0.078\textwidth}{\centering 30 it.}&
\parbox[t]{0.078\textwidth}{\centering 6 it. ($\lambda$)}&
\parbox[t]{0.078\textwidth}{\centering 6 it.}&
\parbox[t]{0.078\textwidth}{\centering 5 it.}&
\parbox[t]{0.078\textwidth}{\centering 30 it.}&
\parbox[t]{0.078\textwidth}{\centering 100 it.}&
\parbox[t]{0.078\textwidth}{\centering 5 it.}\\
\hline
Fig.~\ref{fig-cam4}(a)&
 9.87&13.61 (0.006)&15.37&14.48 (50)&14.48&11.76&14.93&16.86&14.14\\
\hline
Fig.~\ref{fig-cam4g5-6-4i}(a)&
4.67&\phantom07.05 (0.006)&\phantom07.44&\phantom06.60 \phantom0(3)&
\phantom06.59&\phantom05.72&\phantom07.49&\phantom08.17&\phantom07.27\\
\hline
Fig.~\ref{fig-cam4g5-6-4i}(e)&
9.58&11.50 (0.06)\phantom0&11.73&12.59 (25)&12.54&11.42&13.63&13.96&12.92\\
\hline
Fig.~\ref{fig-cam4g5-6-4i}(i)&
3.19&\phantom03.09 (0.16)\phantom0&\phantom01.84&\phantom03.49 \phantom0(1)&
\phantom03.47&\phantom06.46&10.07&13.19&\phantom07.60\\
\hline
\end{tabular}}
\end{table*}

\begin{figure*}[t!]
\centerline{\begin{tabular}{@{}c@{~}c@{~}c@{~}c@{}}
\includegraphics[width=0.24\textwidth]{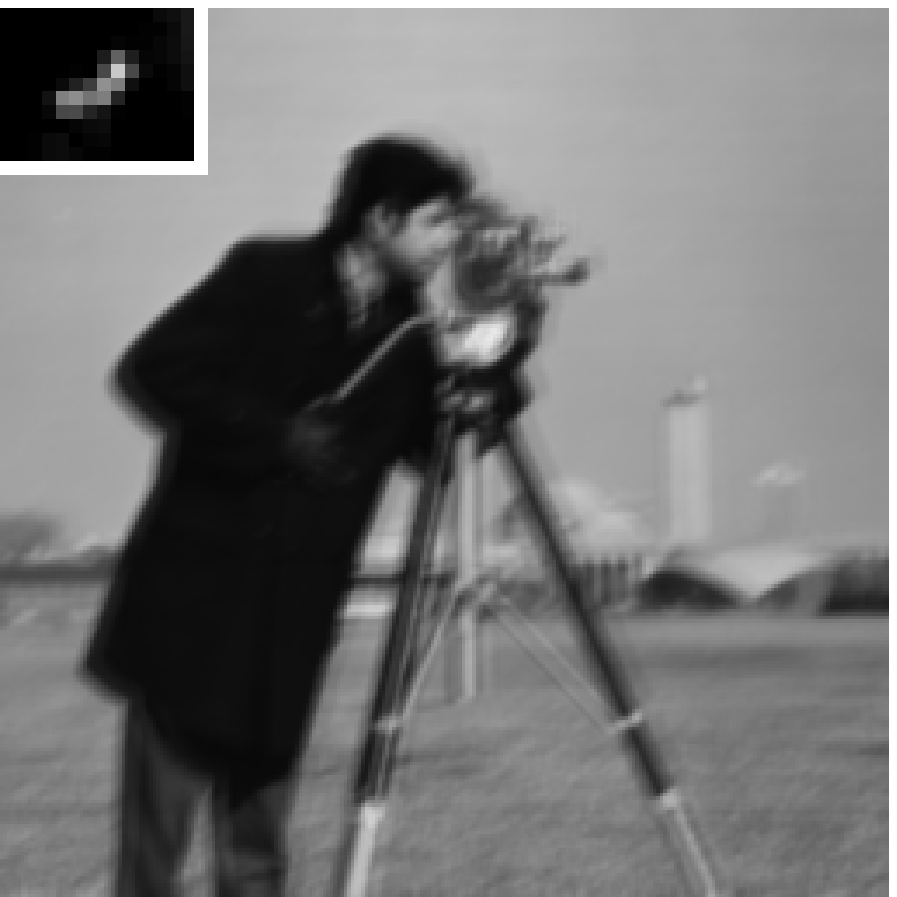}%
 \subfiglabelright{a}&
\includegraphics[width=0.24\textwidth]{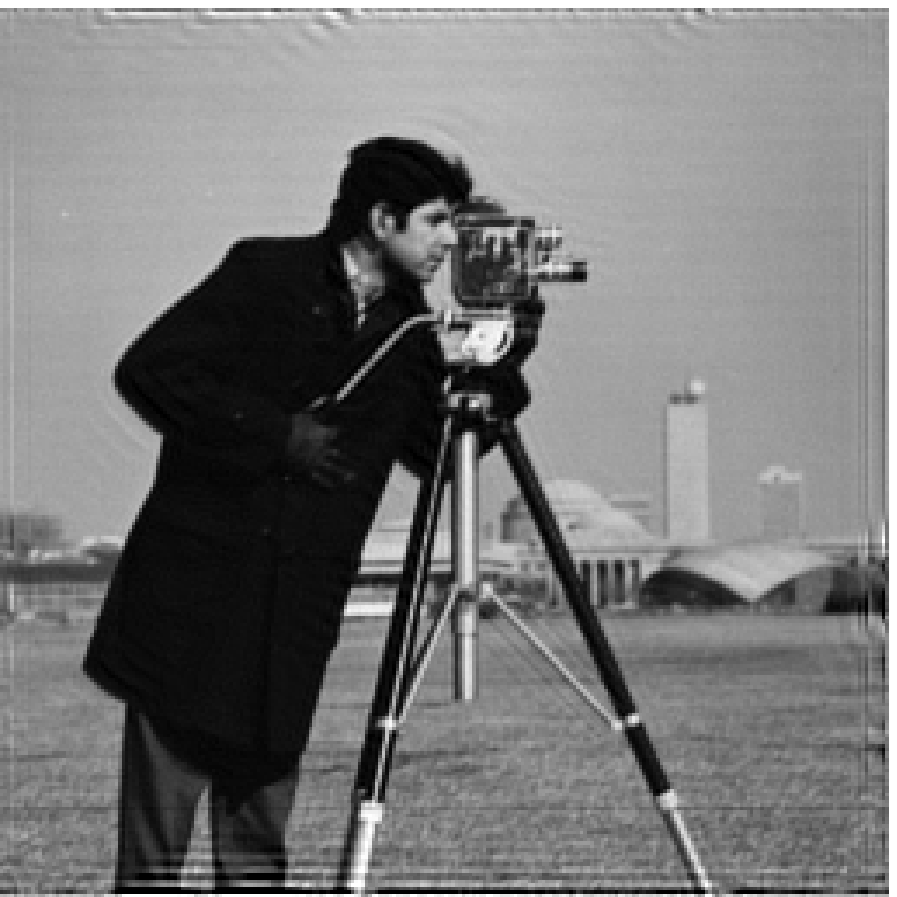}%
 \subfiglabelright{b}&
\includegraphics[width=0.24\textwidth]{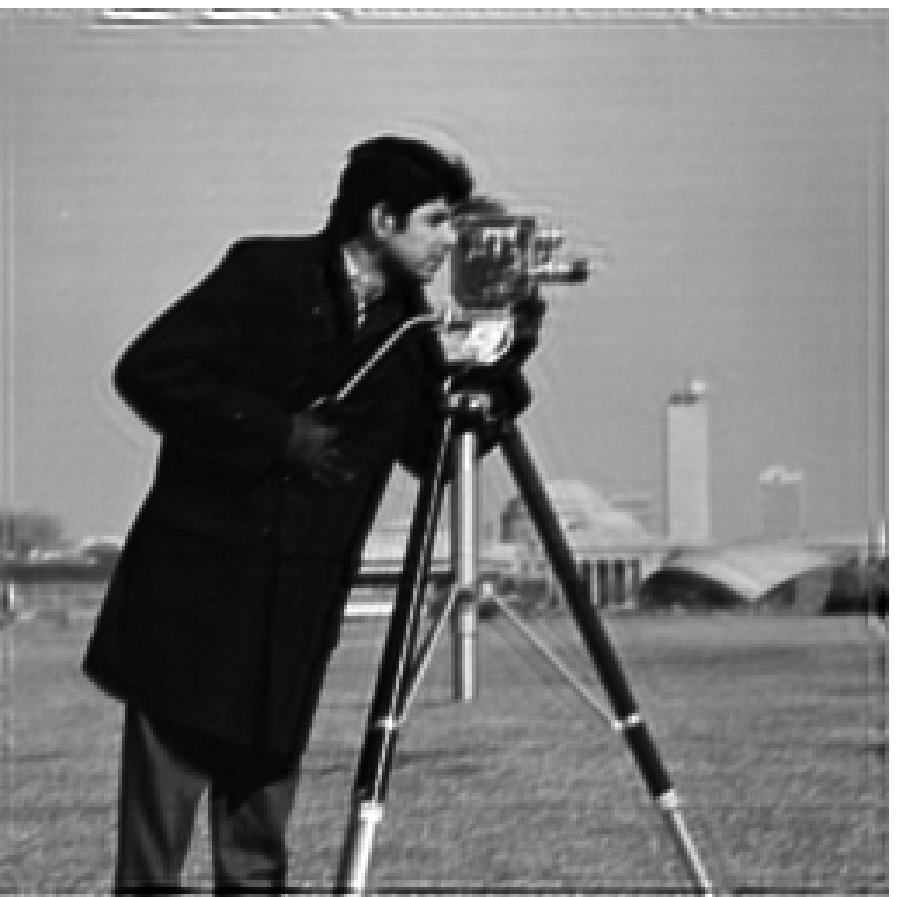}%
 \subfiglabelright{c}&
\includegraphics[width=0.24\textwidth]{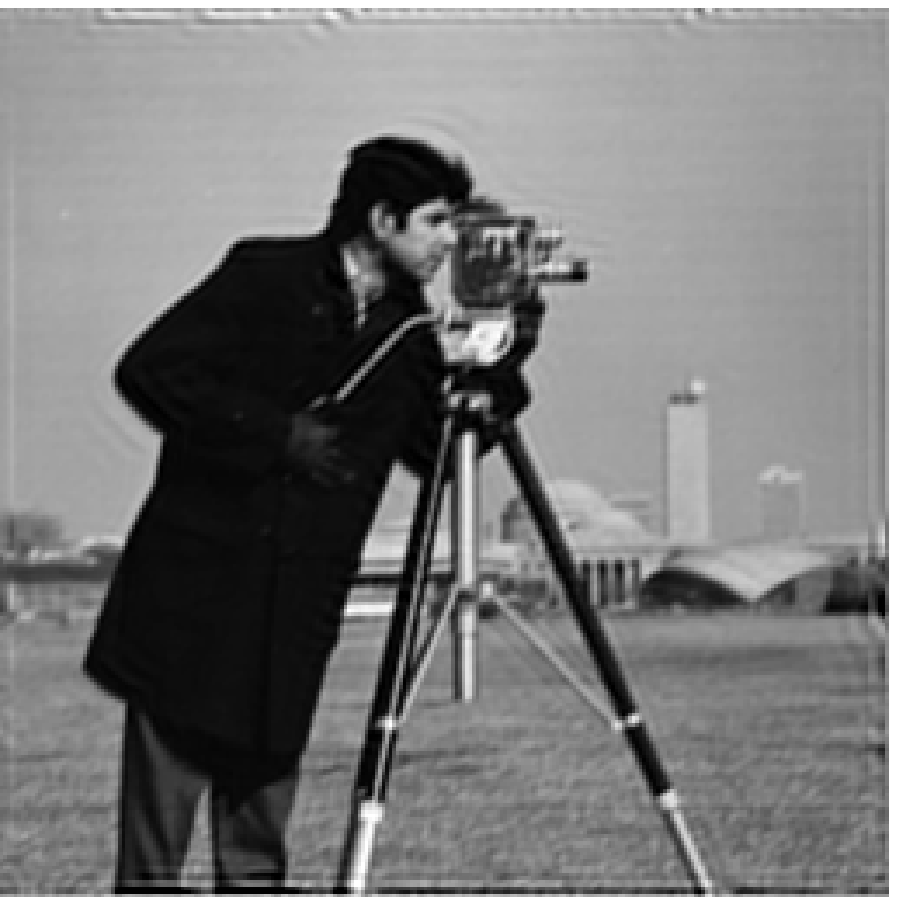}%
 \subfiglabelright{d}\\
\includegraphics[width=0.24\textwidth]{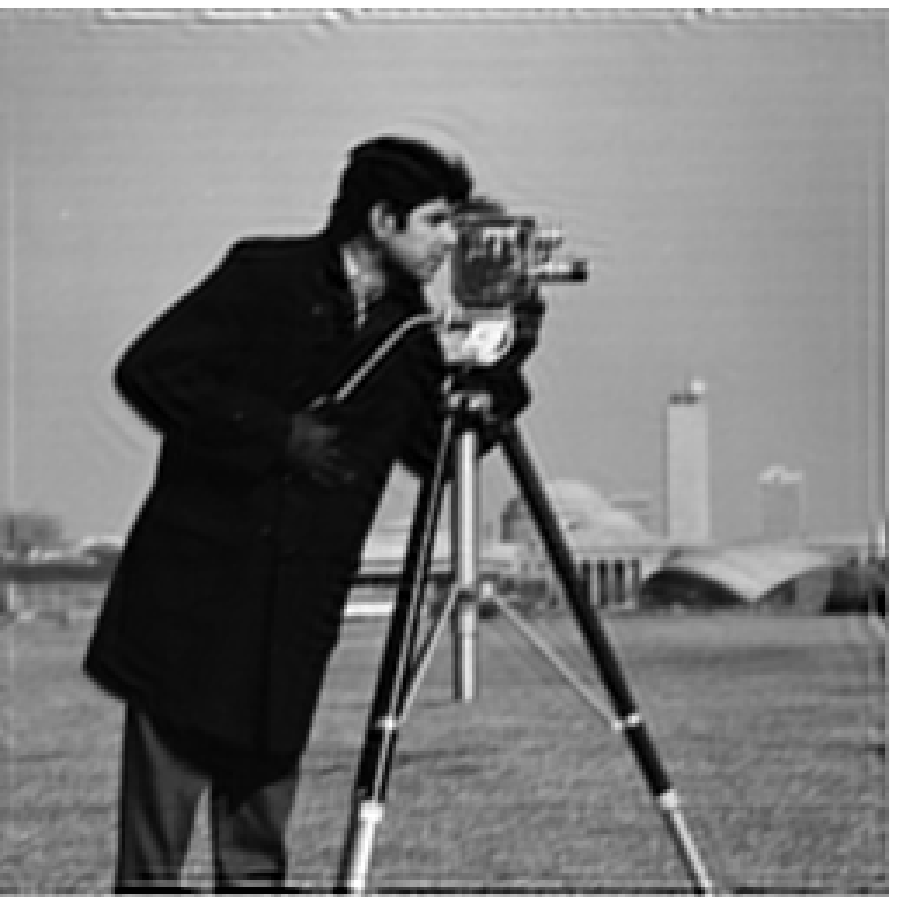}%
 \subfiglabelright{e}&
\includegraphics[width=0.24\textwidth]{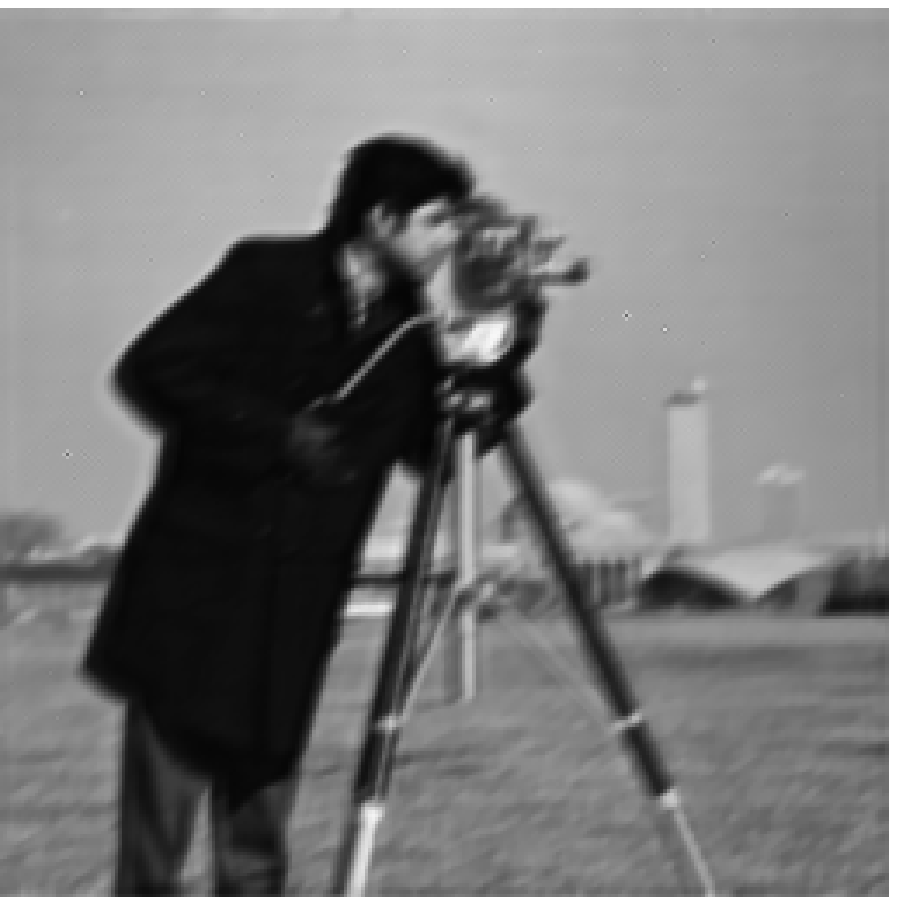}%
 \subfiglabelright{f}&
\includegraphics[width=0.24\textwidth]{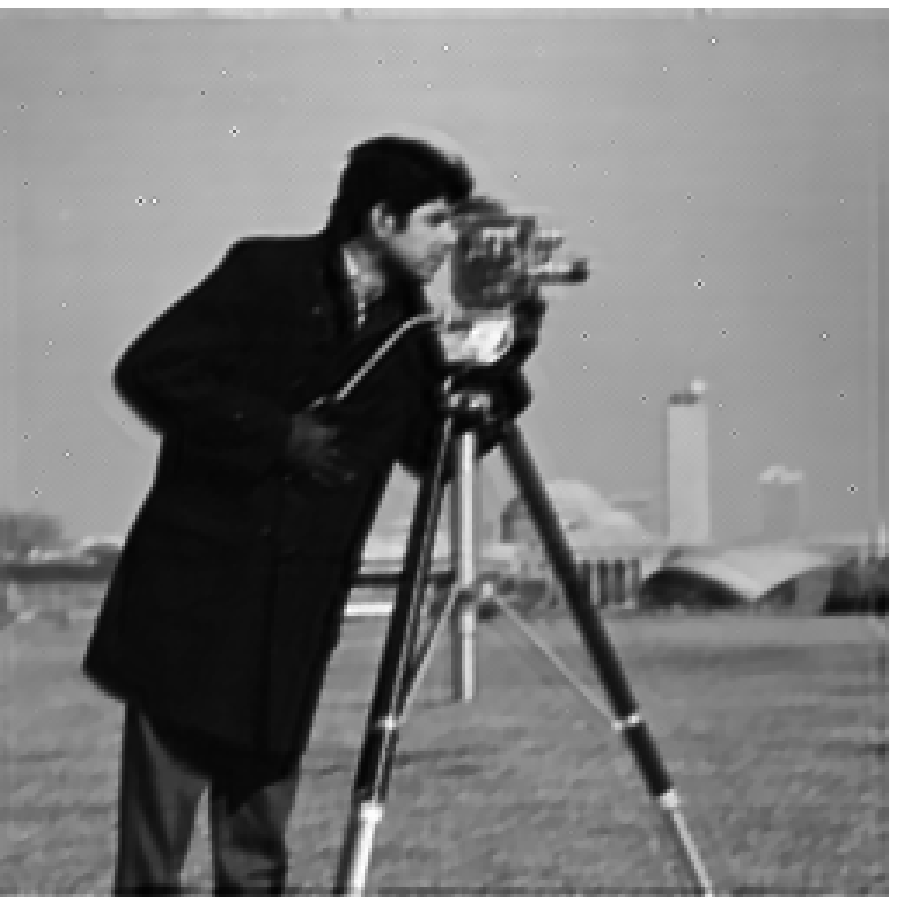}%
 \subfiglabelright{g}&
\includegraphics[width=0.24\textwidth]{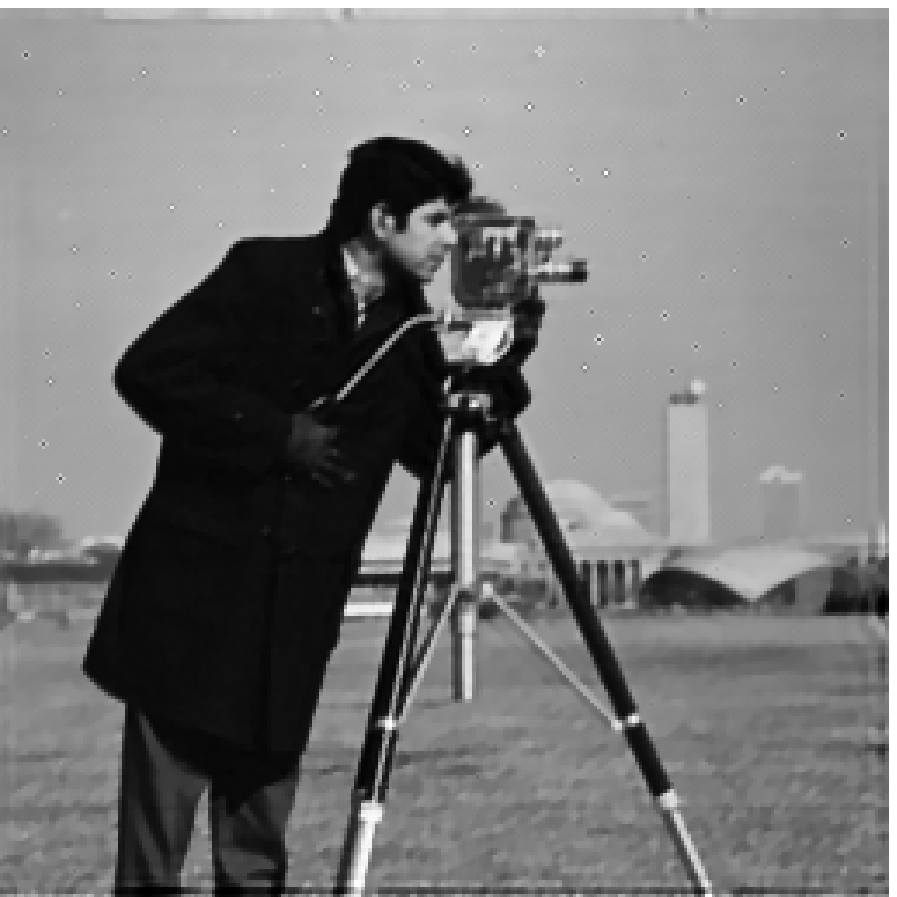}%
 \subfiglabelright{h}
\end{tabular}}
\caption{\label{fig-cam4}%
\textbf{Top row, left to right:}
\textbf{(a)}
\emph{Cameraman} image ($256\times256$ pixels) synthetically
blurred with a small irregular point-spread function (camera-shake type)
using spatial domain convolution with constant continuation. 
Insert shows PSF (four times enlarged). --
\textbf{(b)}
Deblurred by Wiener filtering, $K=0.006$.  --
\textbf{(c)}
Deblurred by 
30 iterations of RL. 
--
\textbf{(d)}
Deblurred by the KF method \cite{Krishnan-nips09} with
$\alpha=2/3$, $\lambda=50$ and $\beta$ from
$1$ to $128\sqrt2$ in powers of $2\sqrt2$, $1$ iteration per level.
\textbf{Bottom row:}
\textbf{(e)}
Deblurred by the WYYZ method \cite{Wang-SIIMS08} with
$\lambda=50$ and $\beta$ from
$1$ to $128\sqrt2$ in powers of $2\sqrt2$, $1$ iteration per level.
--
\textbf{(f)}
5 iterations of RRRL, $\alpha=0.003$, with TV regulariser.
--
\textbf{(g)}
30 iterations of RRRL, same parameters as (f).
--
\textbf{(h)}
100 iterations of RRRL, same parameters as (f).
}
\end{figure*}

\begin{figure*}[t!]
\centerline{\begin{tabular}{@{}c@{~}c@{~}c@{~}c@{}}
\includegraphics[width=0.24\textwidth]{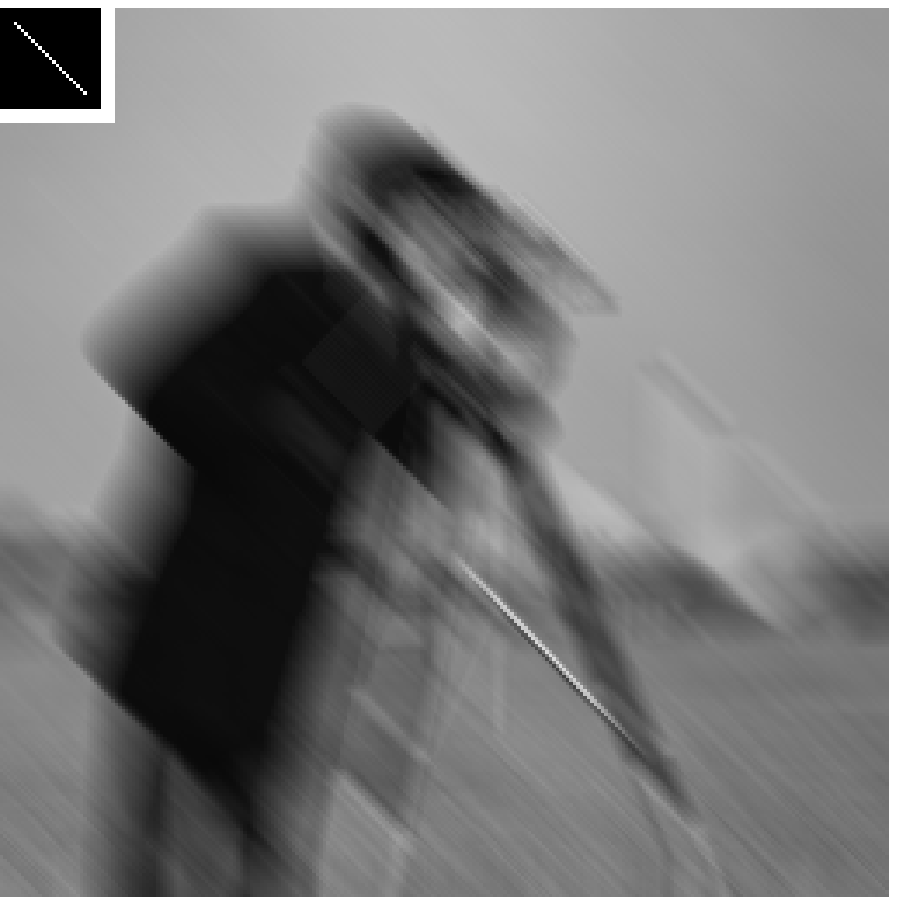}%
 \subfiglabelright{a}&
\includegraphics[width=0.24\textwidth]{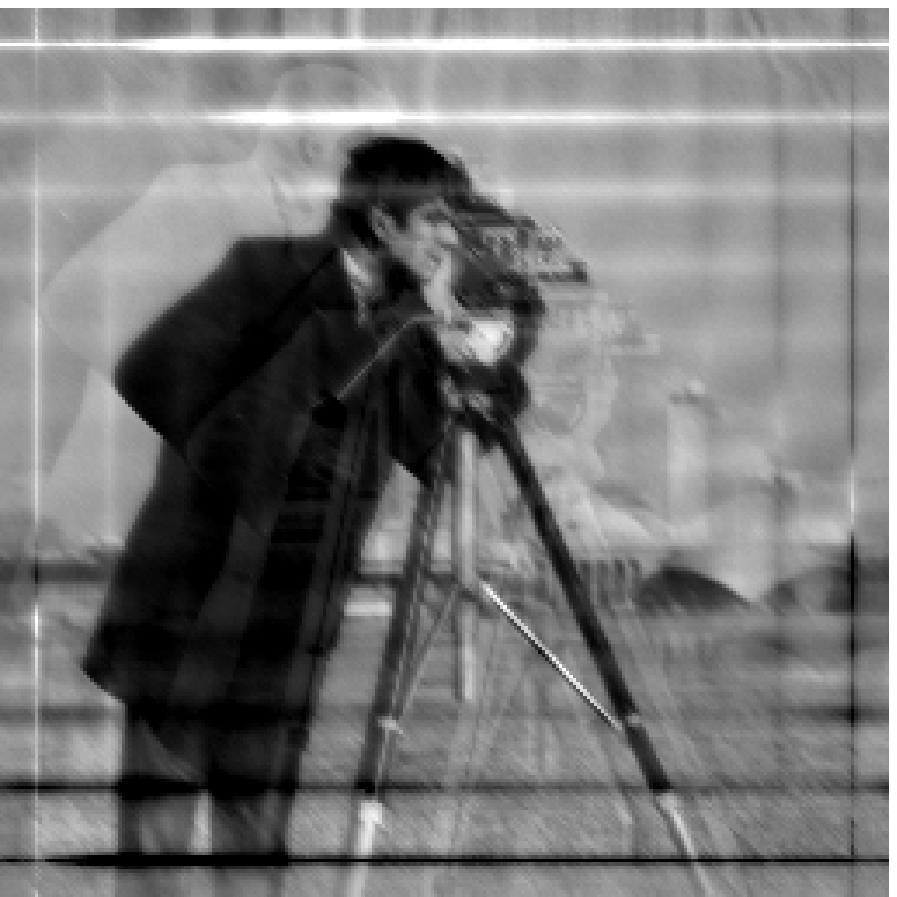}%
 \subfiglabelright{b}&
\includegraphics[width=0.24\textwidth]{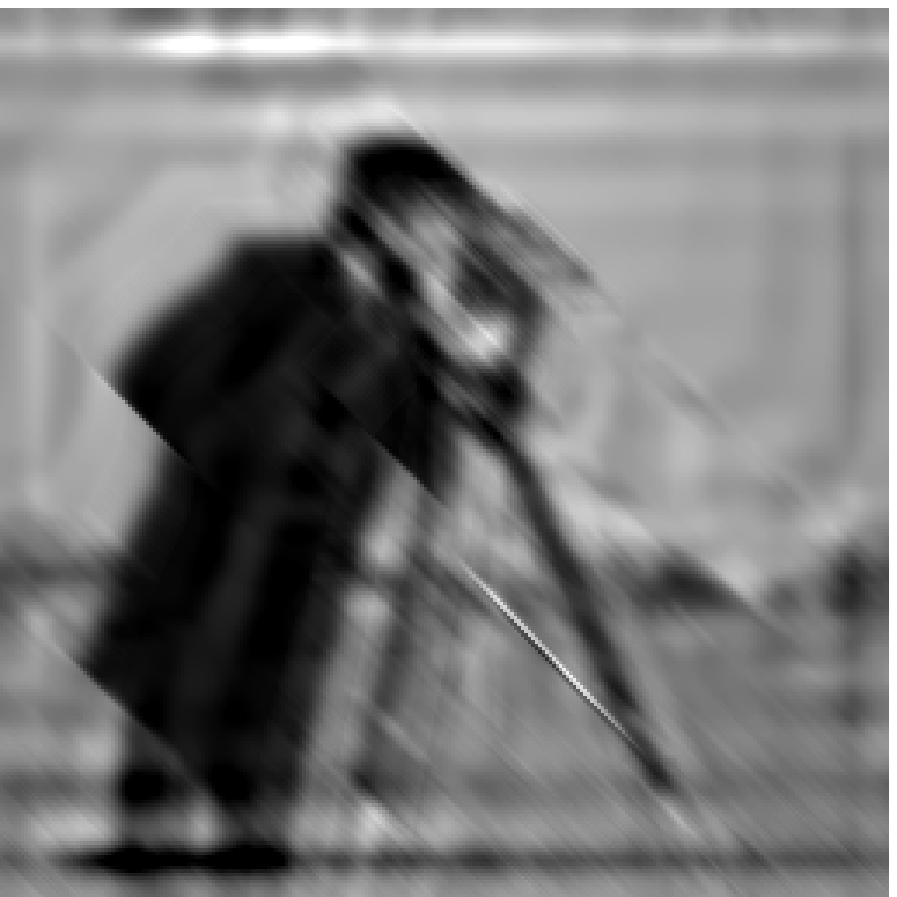}%
 \subfiglabelright{c}&
\includegraphics[width=0.24\textwidth]{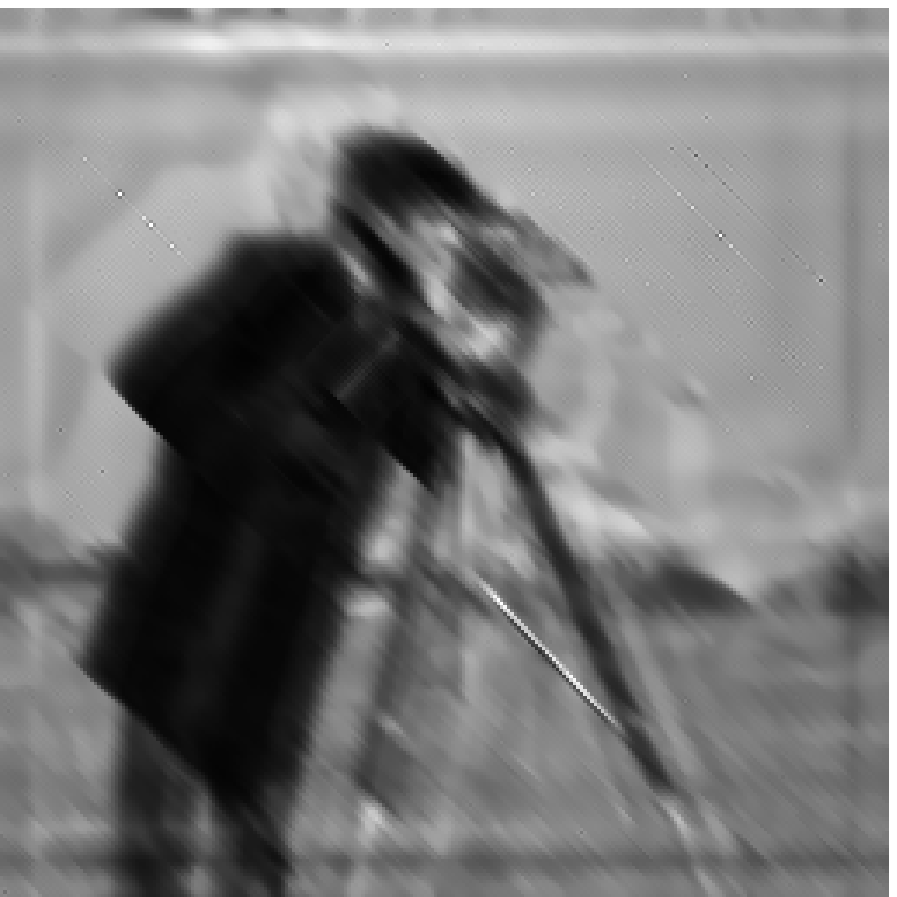}%
 \subfiglabelright{d}\\
\includegraphics[width=0.24\textwidth]{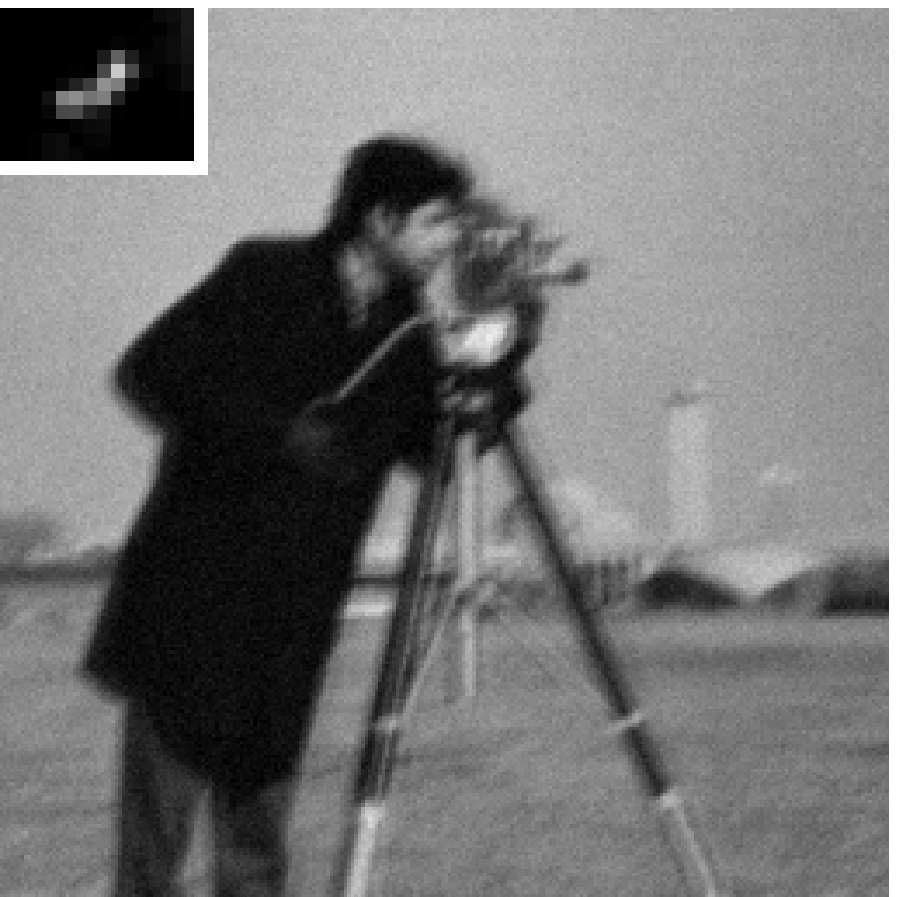}%
 \subfiglabelright{e}&
\includegraphics[width=0.24\textwidth]{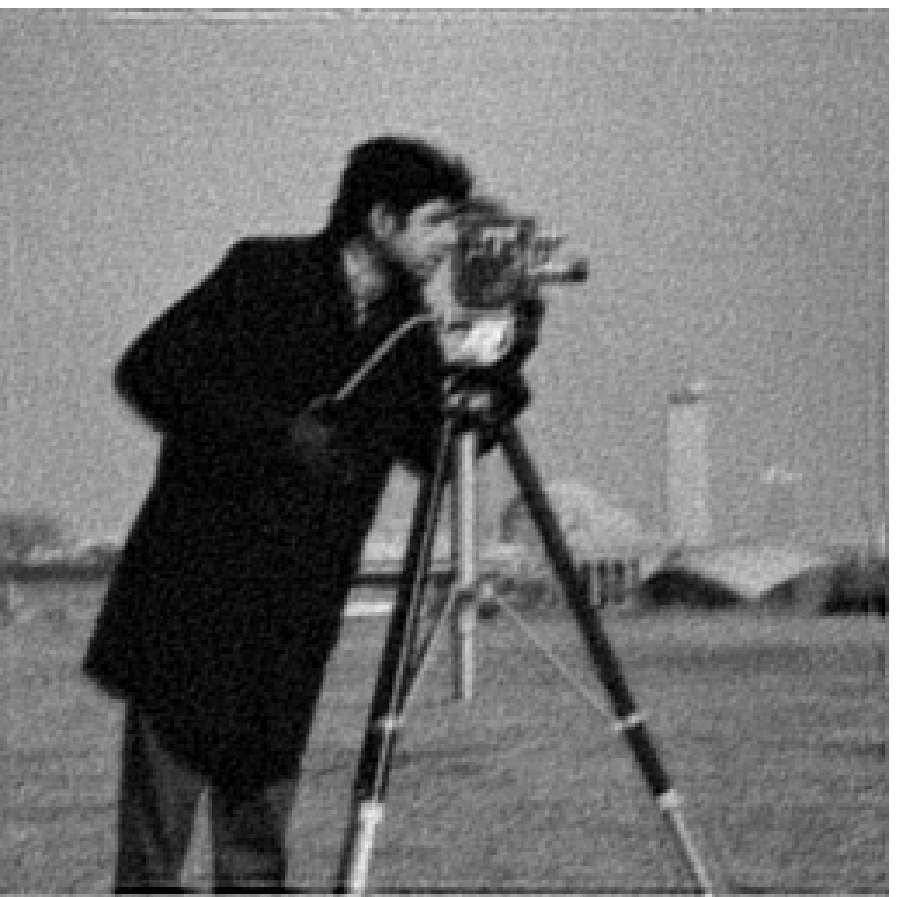}%
 \subfiglabelright{f}&
\includegraphics[width=0.24\textwidth]{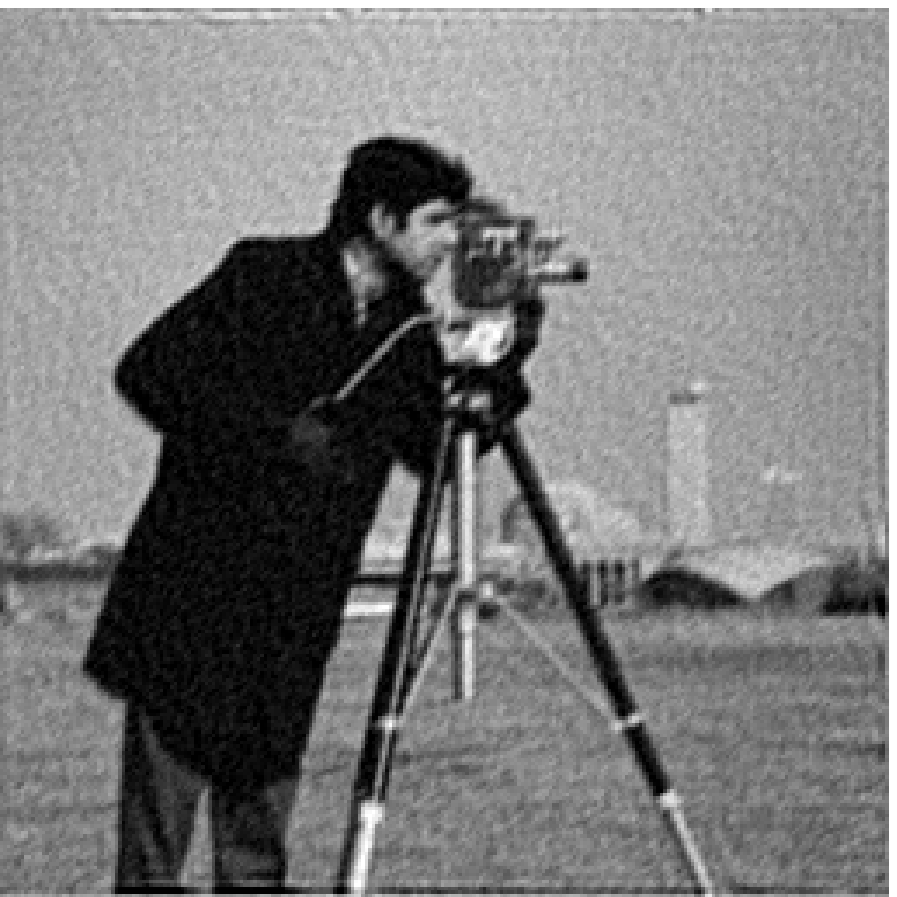}%
 \subfiglabelright{g}&
\includegraphics[width=0.24\textwidth]{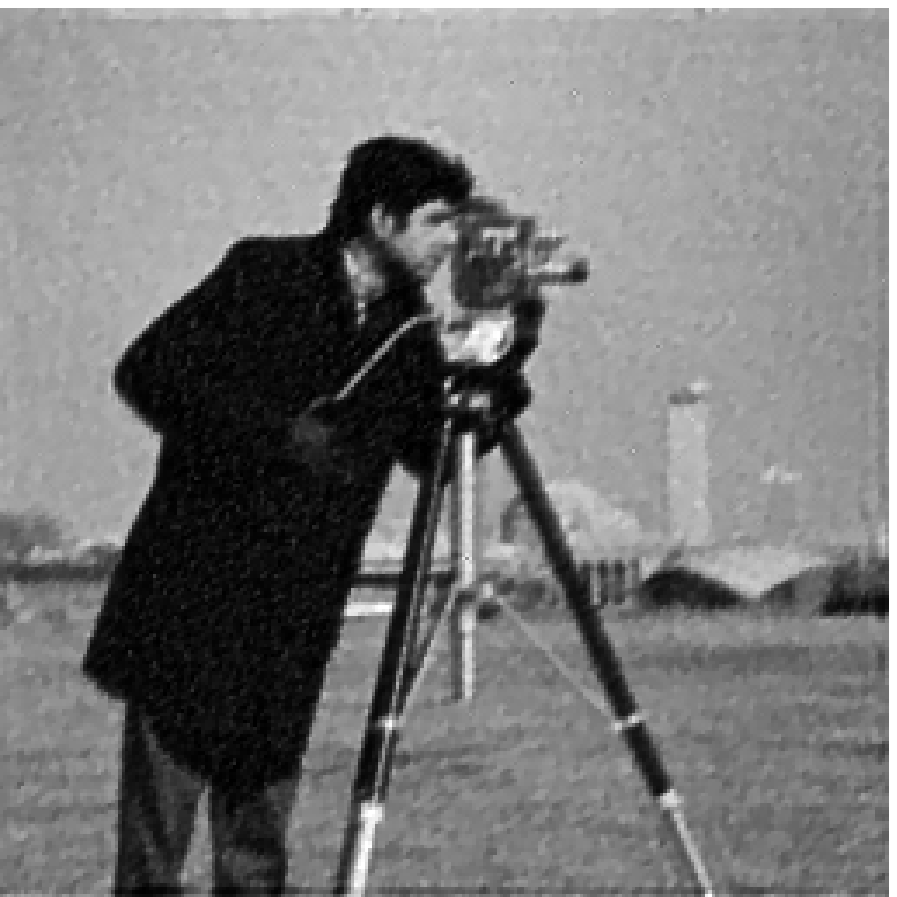}%
 \subfiglabelright{h}\\
\includegraphics[width=0.24\textwidth]{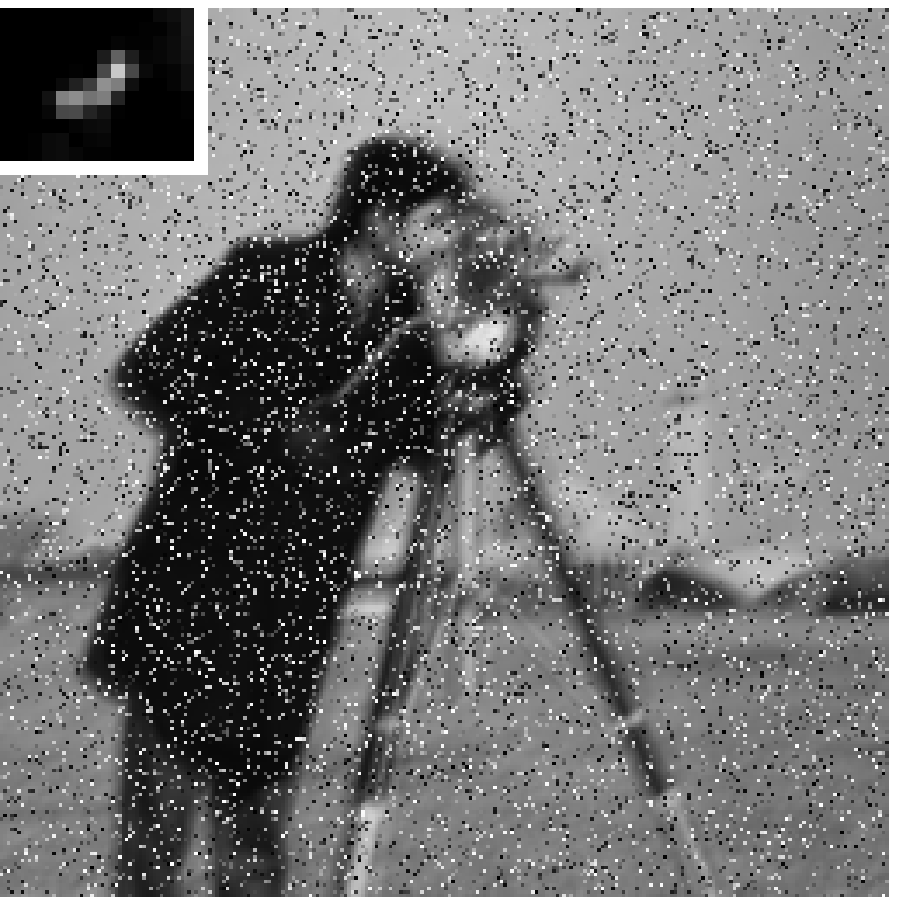}%
 \subfiglabelright{i}&
\includegraphics[width=0.24\textwidth]{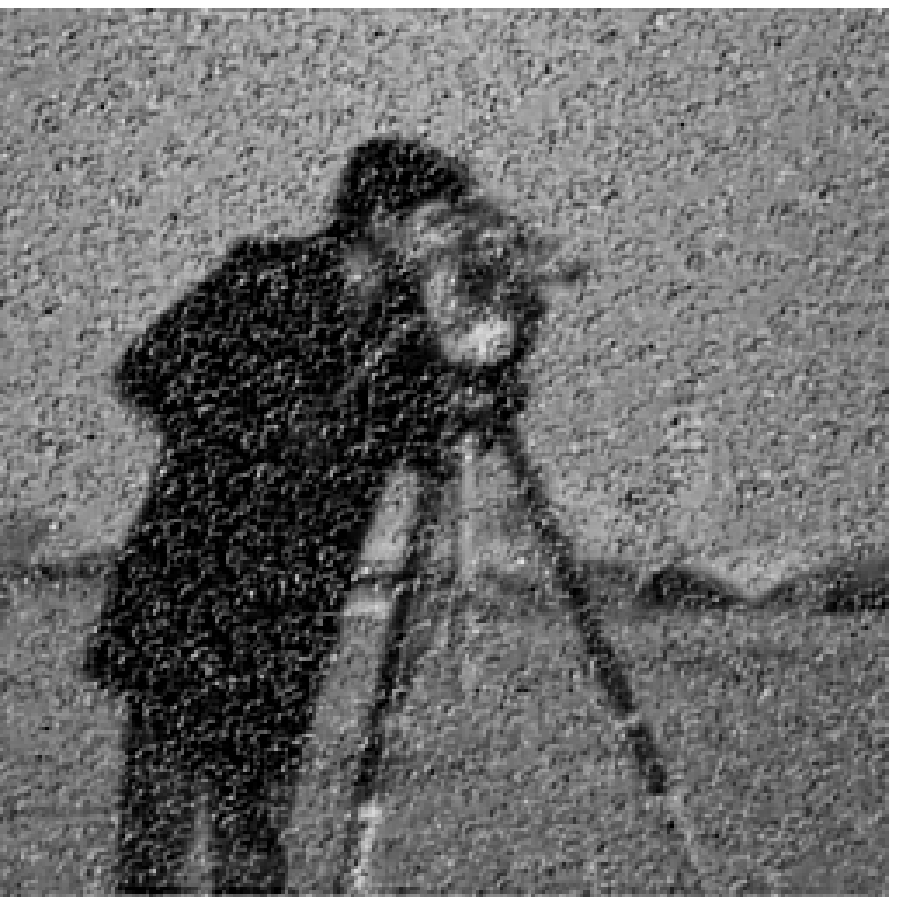}%
 \subfiglabelright{j}&
\includegraphics[width=0.24\textwidth]{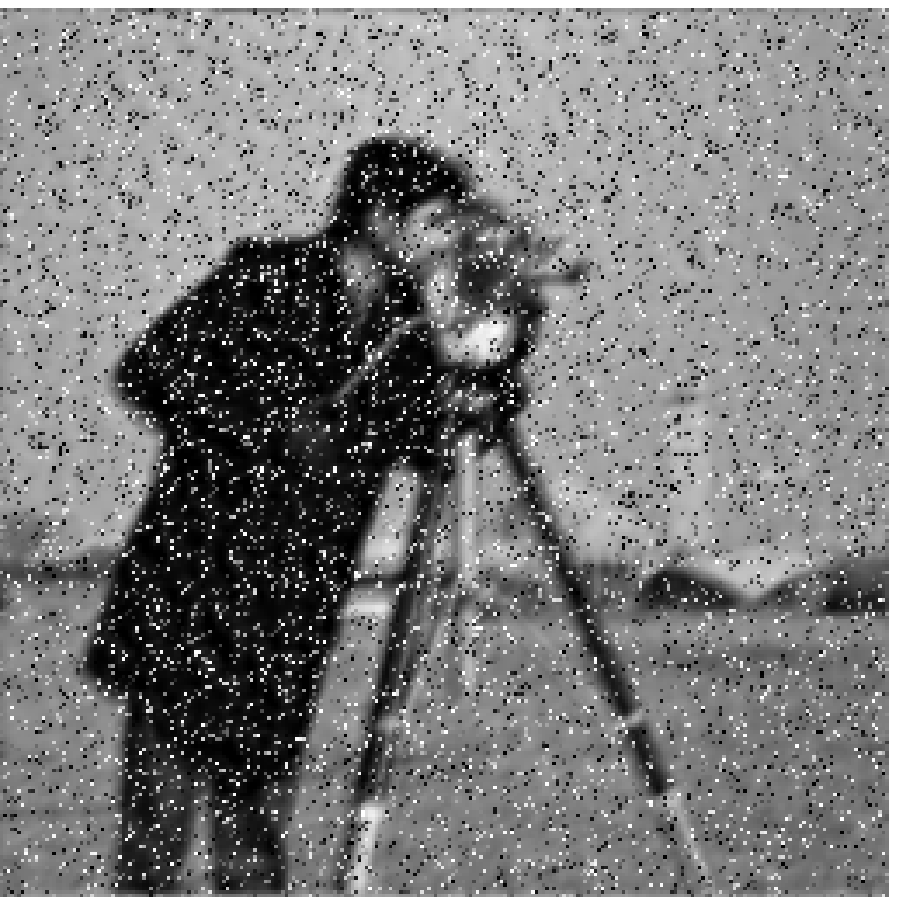}%
 \subfiglabelright{k}&
\includegraphics[width=0.24\textwidth]{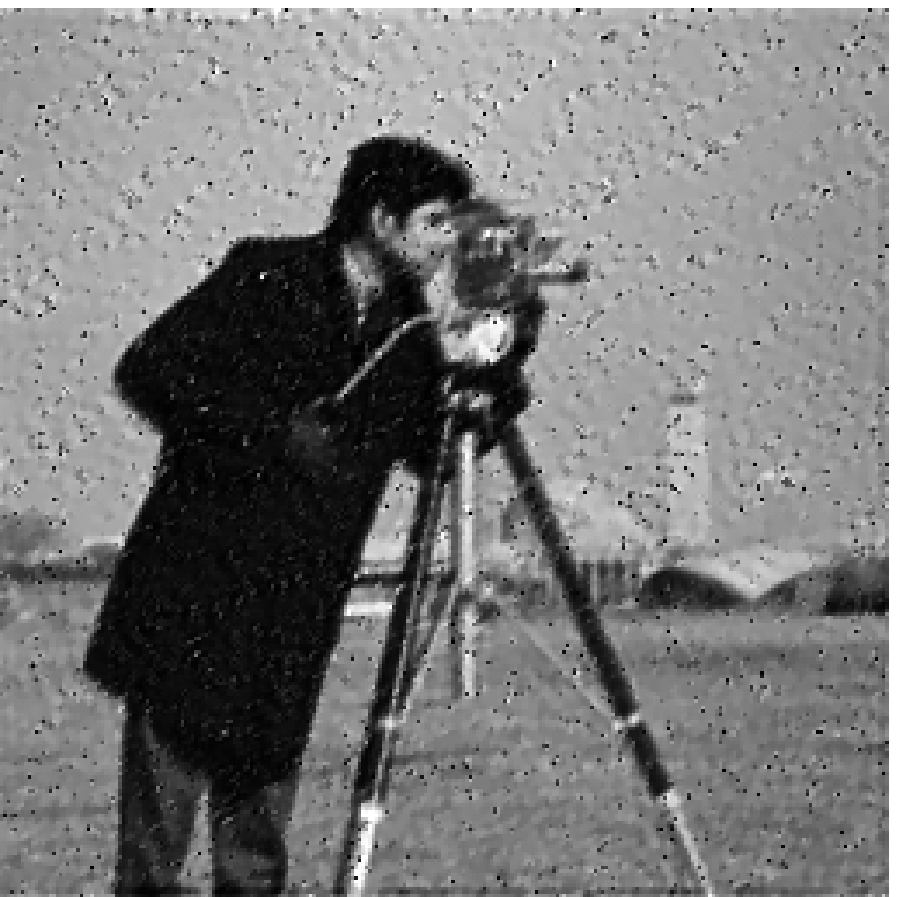}%
 \subfiglabelright{l}
\end{tabular}}
\caption{\label{fig-cam4g5-6-4i}%
\textbf{Top row, left to right:}
\textbf{(a)}
\emph{Cameraman} image synthetically blurred with a motion
blur of 21 pixels length in $45^\circ$ direction
using constant continuation. 
Insert shows PSF (true size). 
--
\textbf{(b)}
Wiener filter, $K=0.006$.
--
\textbf{(c)}
KF method, same parameters as in
Fig.~\ref{fig-cam4}(c) except for $\lambda=3$.
--
\textbf{(d)}
30 iterations of RRRL, same parameters as in
Fig.~\ref{fig-cam4}(h).
\textbf{Middle row, left to right:}
\textbf{(e)} 
Blurred \emph{cameraman} image from Fig.~\ref{fig-cam4}(a)
with additive Gaussian noise, $\sigma=5$. 
Insert shows PSF (four times enlarged). 
--
\textbf{(f)}
Wiener filter, $K=0.06$.
--
\textbf{(g)}
KF method, same parameters as in
Fig.~\ref{fig-cam4}(c) except for $\lambda=25$.
--
\textbf{(h)}
RRRL as in (d).
\textbf{Bottom row, left to right:}
\textbf{(i)}
Blurred \emph{cameraman} image from Fig.~\ref{fig-cam4}(a),
with $15\,\%$ of the pixels replaced by noise pixels with
uniform distribution on $[0,255]$.
Insert shows PSF (four times enlarged). 
--
\textbf{(j)}
Wiener filter, $K=0.16$.
--
\textbf{(k)}
KF method, same parameters as in
Fig.~\ref{fig-cam4}(c) except for $\lambda=1$.
--
\textbf{(l)}
RRRL as in (d).
}
\end{figure*}

We compare thus Wiener filter, Richardson-Lucy
deconvolution (RL), the iterative methods by
Wang et al. (WYYZ) \cite{Wang-SIIMS08} and by
Krishnan and Fergus (KF) \cite{Krishnan-nips09}, 
and RRRL \cite{Elhayek-dagm11,Welk-tr10}.
{\sloppy\par}

\subsection{Synthetically Blurred Images}
Our first test scenario is based on a version 
of the popular \emph{cameraman} test image which has been 
synthetically blurred with a space-invariant ``camera-shake''-like
point-spread function of irregular shape, Fig.~\ref{fig-cam4}(a).
No noise besides the quantisation noise is present in this
test image.

Since the deconvolution methods are implemented with Fourier
domain convolution, the blur in the test images has been carried 
out via the spatial domain in order
to avoid what is known as an ``inverse crime'' \cite{Colton-Book92}.
The boundary condition in the convolution was chosen as constant
continuation (along normals to the boundary) which contrasts
to the periodic continuation implicitly involved in the Fourier
convolution in the deblurring algorithms.

Table~\ref{tab-snr} collects signal-to-noise ratios 
\[
\mathrm{SNR}(u,u_0) = 
10\,\log_{10}\frac{\mathrm{var}\,(u)}{\mathrm{var}\,(u-u_0)}
\,\mathrm{dB}
\]
for blurred and deblurred images $u$ compared to the unperturbed 
cameraman image $u_0$.

In Fig.~\ref{fig-cam4g5-6-4i}, we test Wiener filter, KF and 
RRRL on three other synthetically blurred \emph{cameraman} images 
to study the
stability of results under additional noise and stronger blur.
For signal-to-noise ratios see again Table~\ref{tab-snr}.
In Fig.~\ref{fig-cam4g5-6-4i}(a--d), a fairly low amount of Gaussian
noise is added to the test image. Subfig.~(e--h) consider
a spatially more extended point-spread function. 
A more drastic noise -- impulse noise with $15\,\%$ density --
is added to the first test image in Subfig.~(i--l).

It is evident both visually and from the SNR figures that RRRL, KF,
and WYYZ generally allow for a good restoration. The Wiener filter
is sensitive to boundary artifacts for larger blurs, and to non-Gaussian
noise. The robust data term of the RRRL model gives it also an advantage
over the KF and WYYZ methods in settings with boundary artifacts and
more severe noise.

\subsection{Combined Wiener-RRRL Method (WR\textsuperscript3L)}

It should be noted that in spite of its favourable properties RRRL 
still takes fairly many iterations (30\ldots100, depending
on noise level) to achieve an acceptable degree of sharpness. 
On the other hand, Wiener filtering, being a non-iterative method, 
provides a reasonable sharpness in one fast computation step
but at the cost that the remaining noise and ringing artifacts
are more pronounced. 

This motivates us to test a combined approach in which the
Wiener filter is used as a first step, followed by an RRRL
iteration for which the Wiener result acts as initialisation.
A caveat in doing so is that the Wiener filter output can contain
zero or negative grey-values, which cannot be handled within the
RRRL model. However, negative grey-values appear as part of 
artifacts anyway, so this can be countered simply by replacing all
zero or negative grey-values to a small positive value in the 
input for RRRL.

Table~\ref{tab-snr} includes this method, WR\textsuperscript3L,
in the last column. Note that even with as few as 5 iterations
WR\textsuperscript3L performs in most cases comparable to about
30 iterations of pure RRRL. Visual evaluation of 
WR\textsuperscript3L will be included in subsequent experiments.

\begin{figure*}[t!]
\centerline{\begin{tabular}{@{}c@{~}c@{~}c@{~}c@{}}
\includegraphics[width=0.24\textwidth]{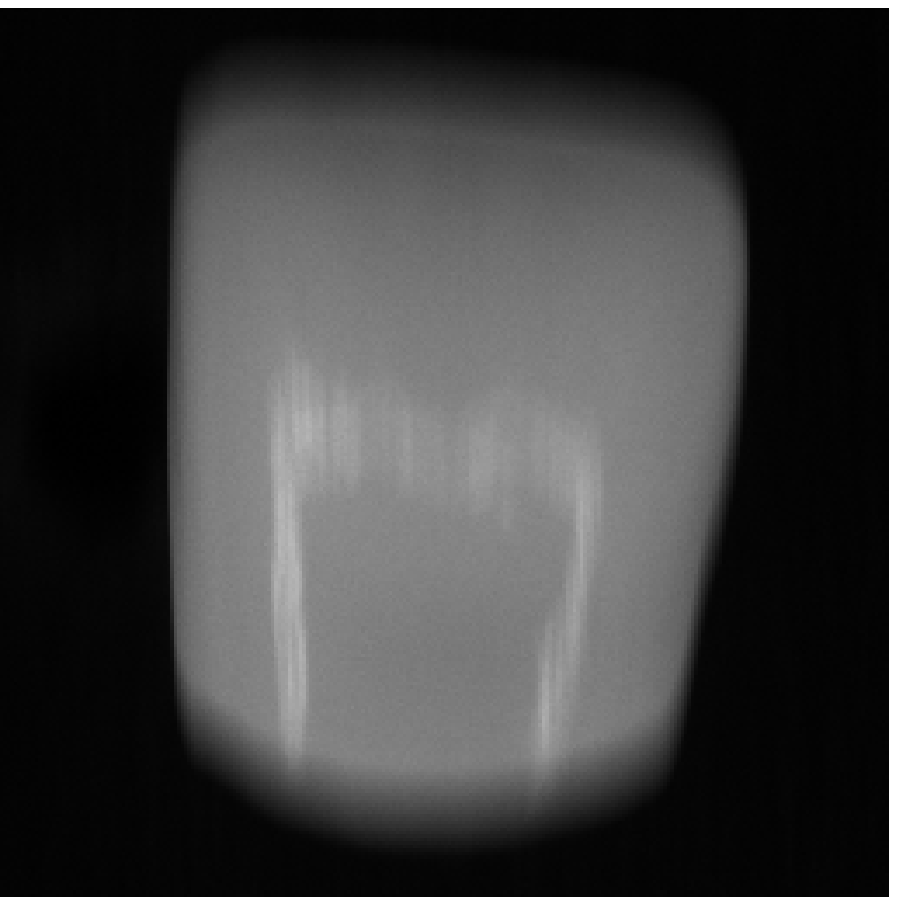}%
 \subfiglabelright{a}&
\includegraphics[width=0.24\textwidth]{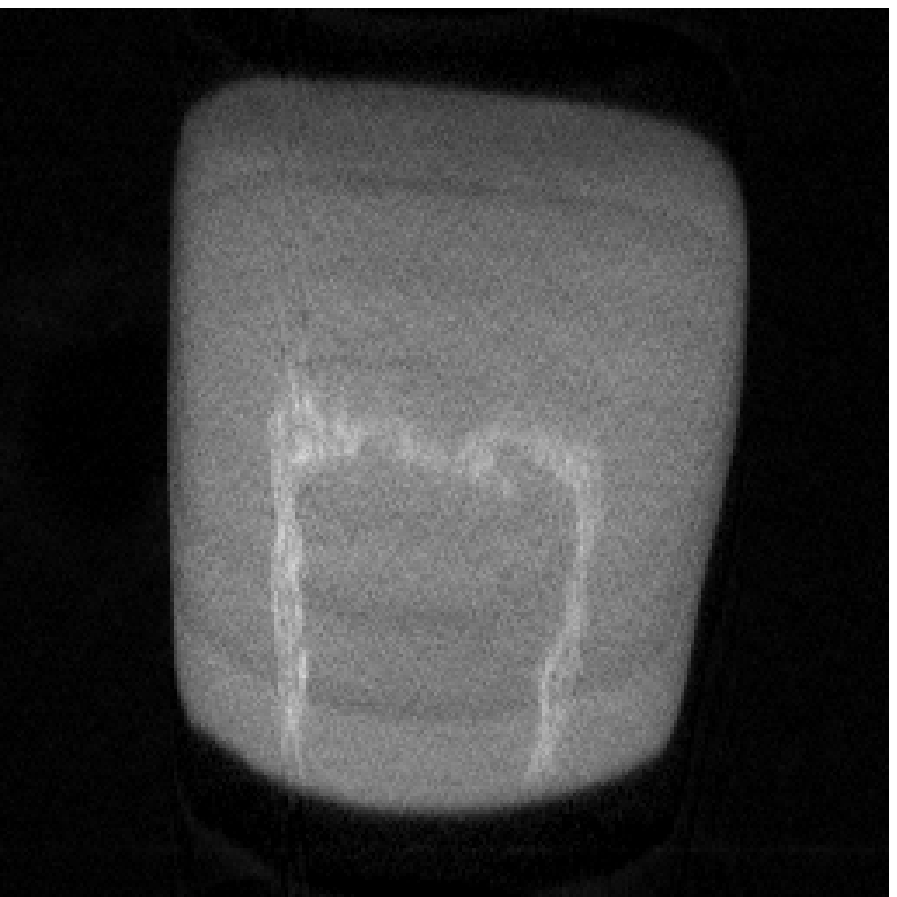}%
 \subfiglabelright{b}&
\includegraphics[width=0.24\textwidth]{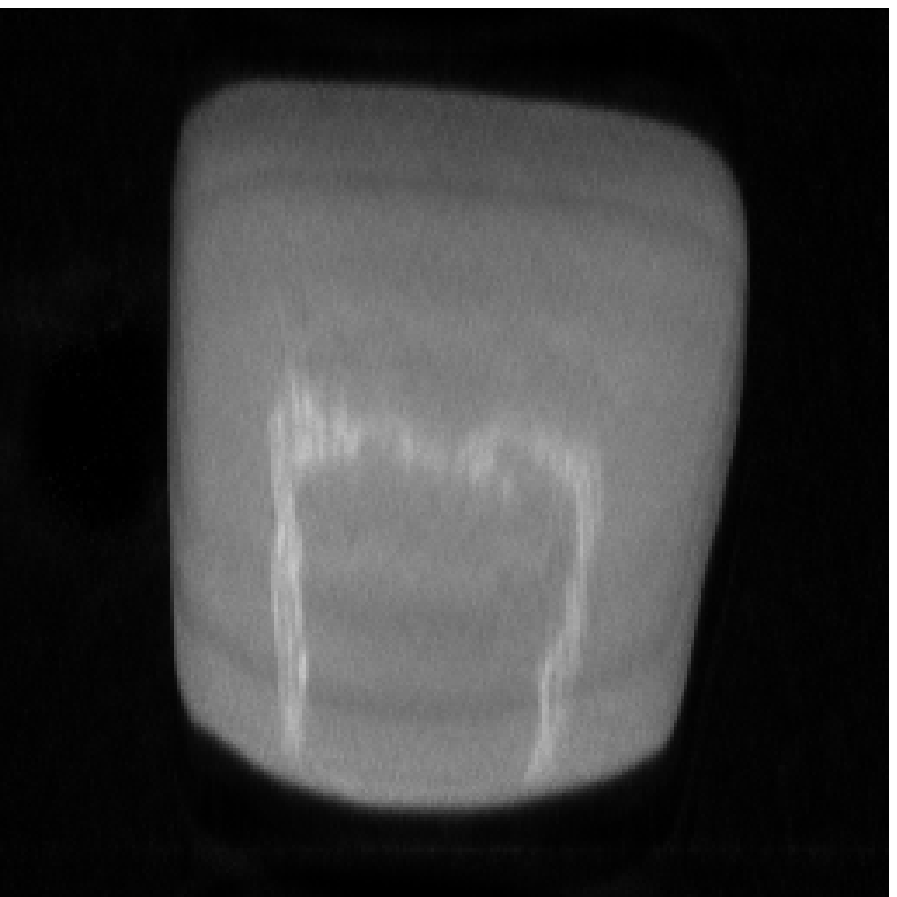}%
 \subfiglabelright{c}&
\includegraphics[width=0.24\textwidth]{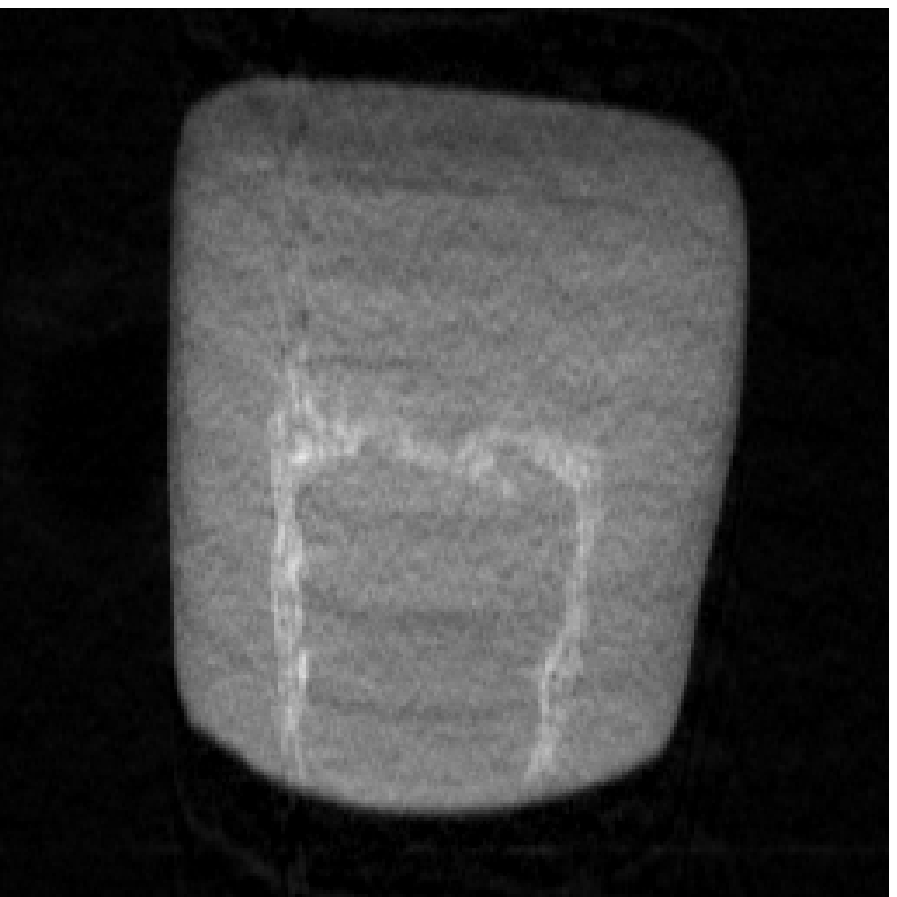}%
 \subfiglabelright{d}\\
\includegraphics[width=0.24\textwidth]{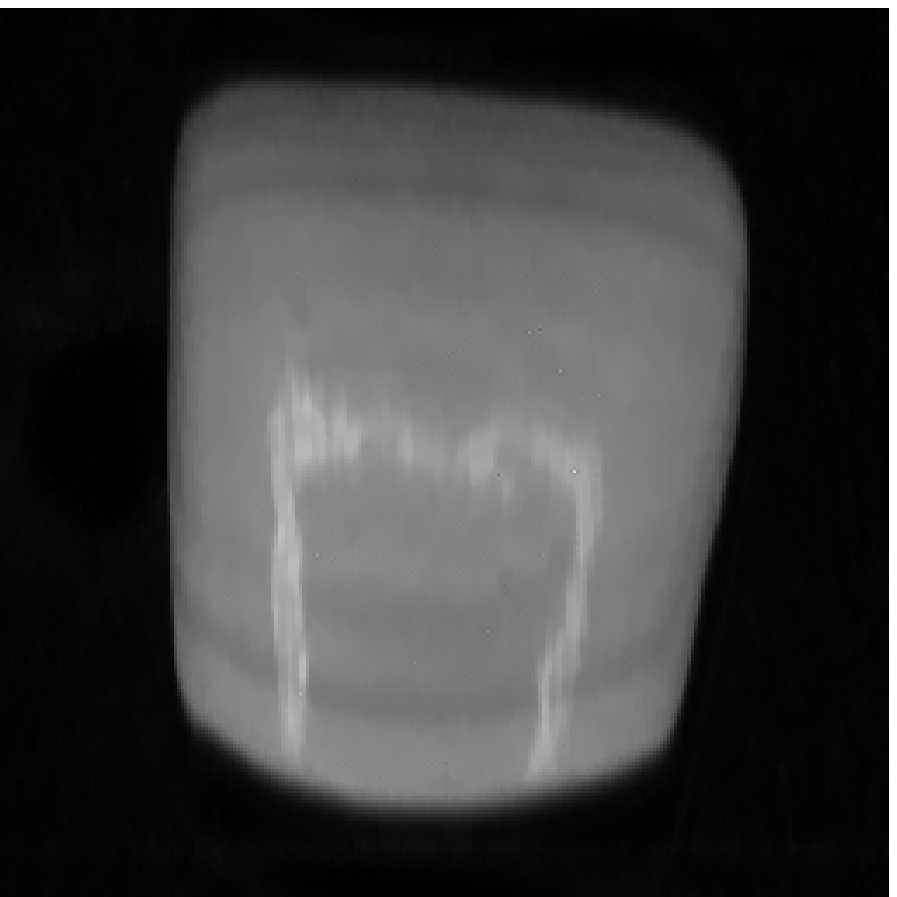}%
 \subfiglabelright{e}&
\includegraphics[width=0.24\textwidth]{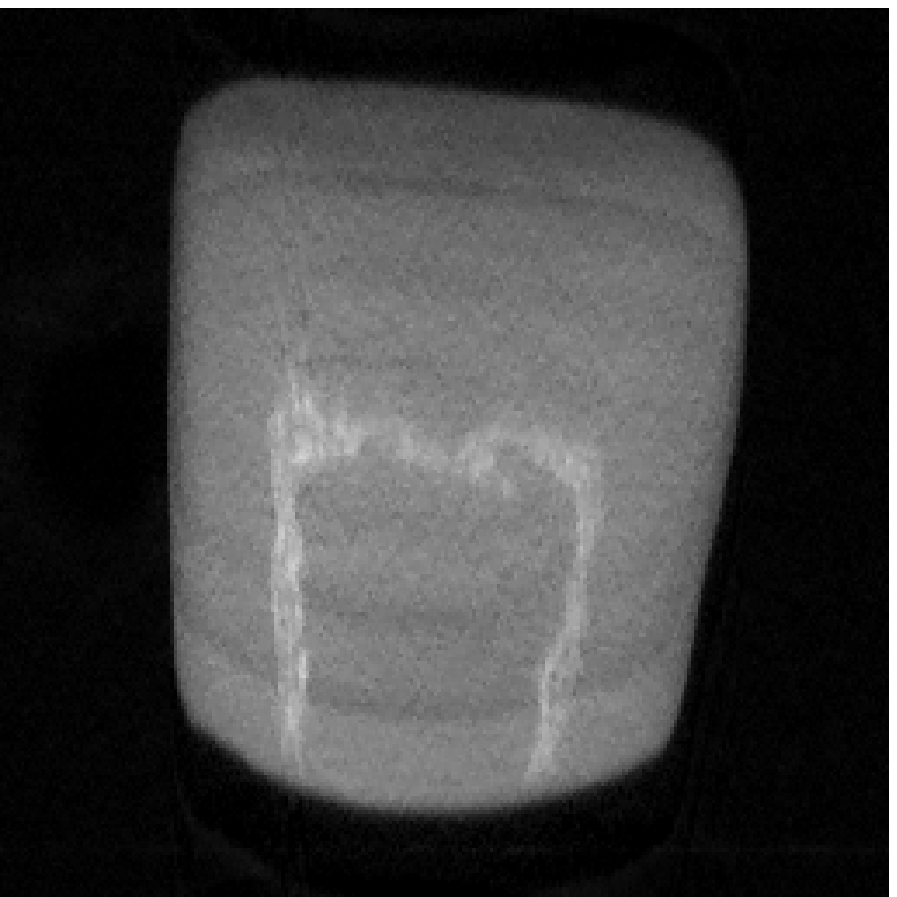}%
 \subfiglabelright{f}&
\includegraphics[width=0.24\textwidth]{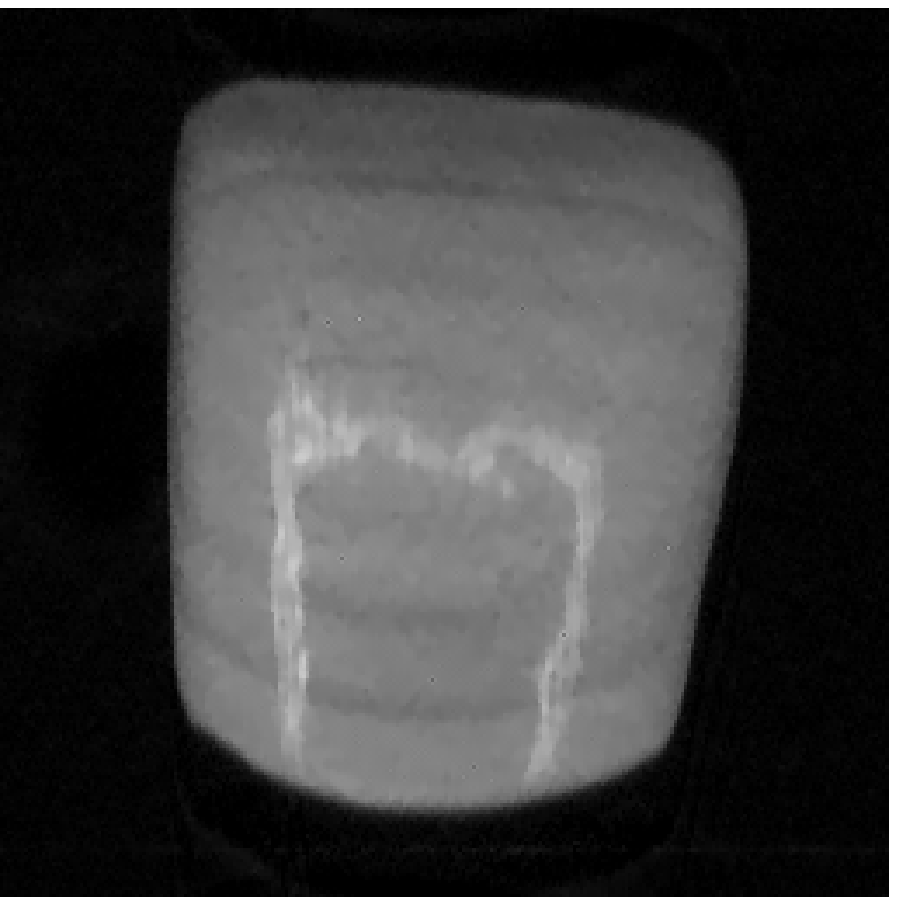}%
 \subfiglabelright{g}&
\includegraphics[width=0.24\textwidth]{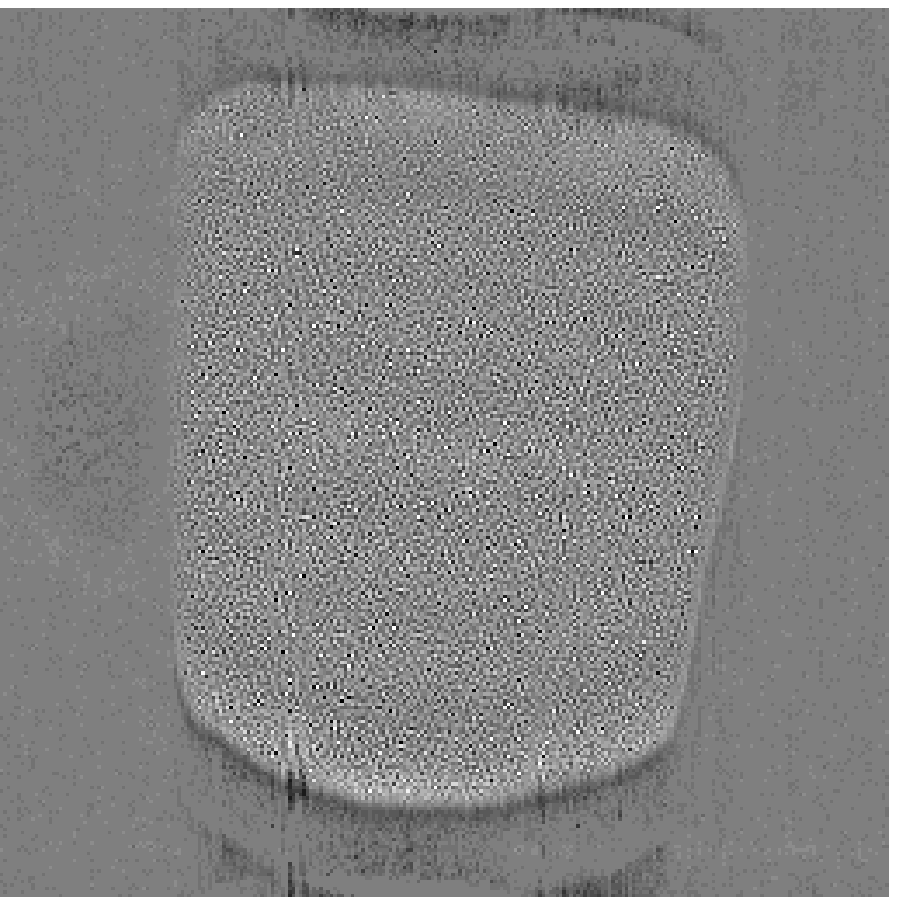}%
 \subfiglabelright{h}
\end{tabular}}
\caption{\label{fig-tooth}
\textbf{Top row, left to right:}
\textbf{(a)}
Original image ($256\times256$ pixels) of a prosthetic 
tooth with motion blur (blur length: 27 pixels). 
Courtesy of Westcam Projektmanagement GmbH. --
\textbf{(b)}
Filtered by Wiener filter, $K=0.006$. --
\textbf{(c)}
Filtered by 30 iterations of RL. --
\textbf{(d)}
Filtered by KF method, $\lambda=200$, $\beta$ from $1$ to $128\sqrt2$ in
multiplicative steps of $2\sqrt2$, $1$ iteration per level.
\textbf{Bottom row, left to right:}
\textbf{(e)}
Filtered by 30 iterations of RRRL with TV regulariser, 
$\alpha=0.003$. --
\textbf{(f)}
WR\textsuperscript3L: Wiener filtering and 5 iterations of
RRRL with TV regulariser, $\alpha=0.003$. --
\textbf{(g)}
Same but 30 iterations. --
\textbf{(h)} 
Difference image of (b) and (f), range $[-10,10]$
rescaled to $[0,255]$.
}
\end{figure*}

\begin{figure*}[t!]
\centerline{\begin{tabular}{@{}c@{~}c@{~}c@{~}c@{}}
\includegraphics[width=0.24\textwidth]{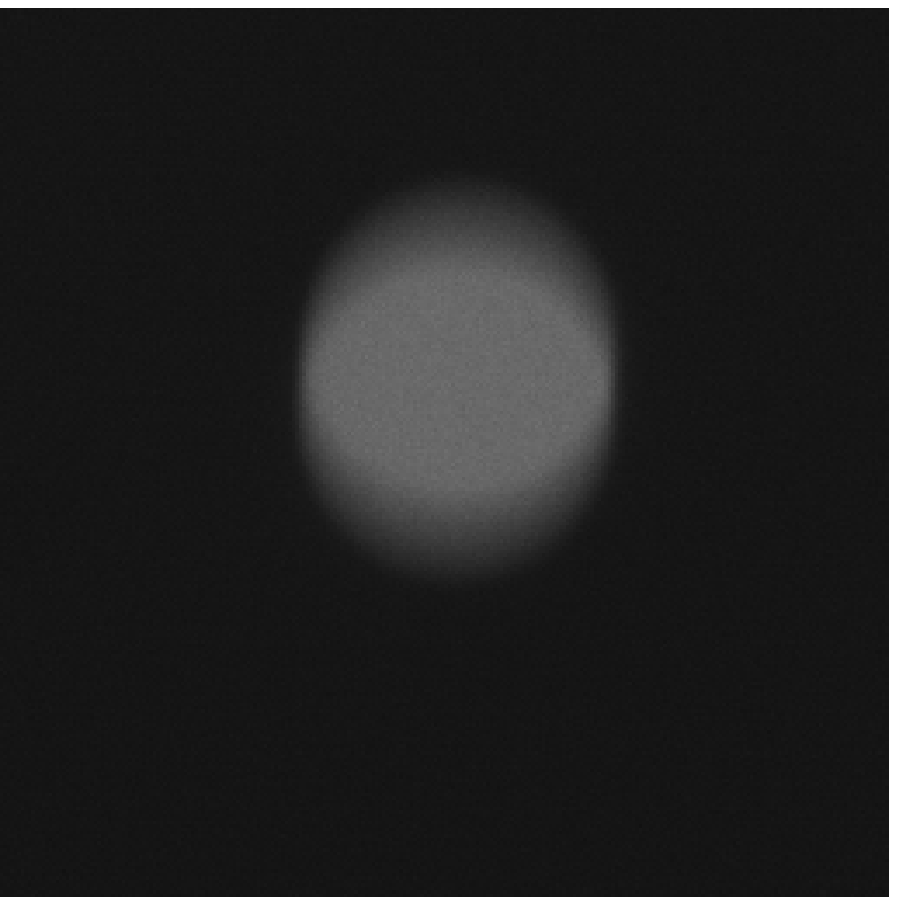}%
 \subfiglabelright{a}&
\includegraphics[width=0.24\textwidth]{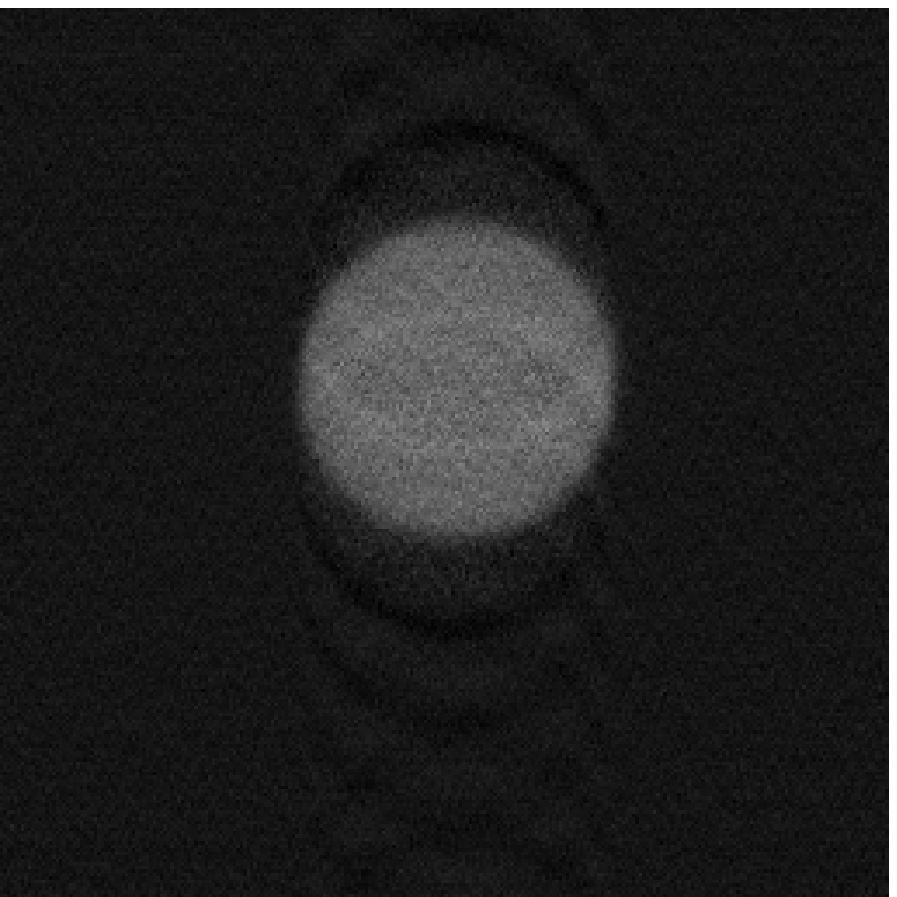}%
 \subfiglabelright{b}&
\includegraphics[width=0.24\textwidth]{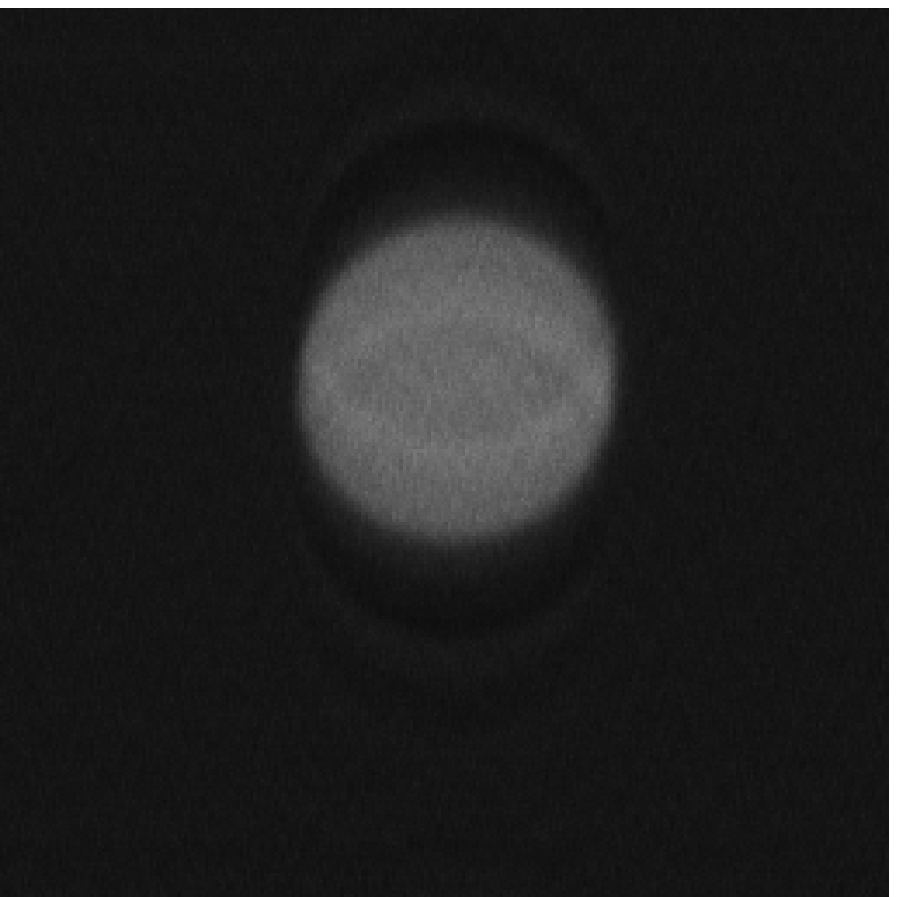}%
 \subfiglabelright{c}&
\includegraphics[width=0.24\textwidth]{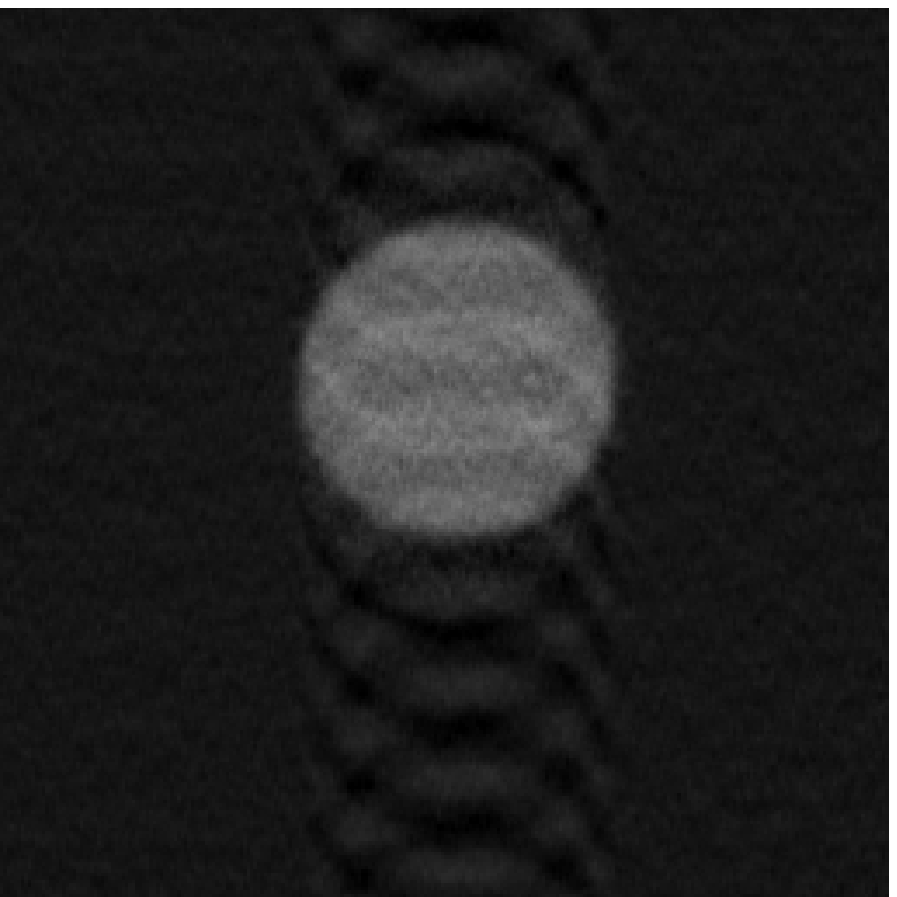}%
 \subfiglabelright{d}\\
\includegraphics[width=0.24\textwidth]{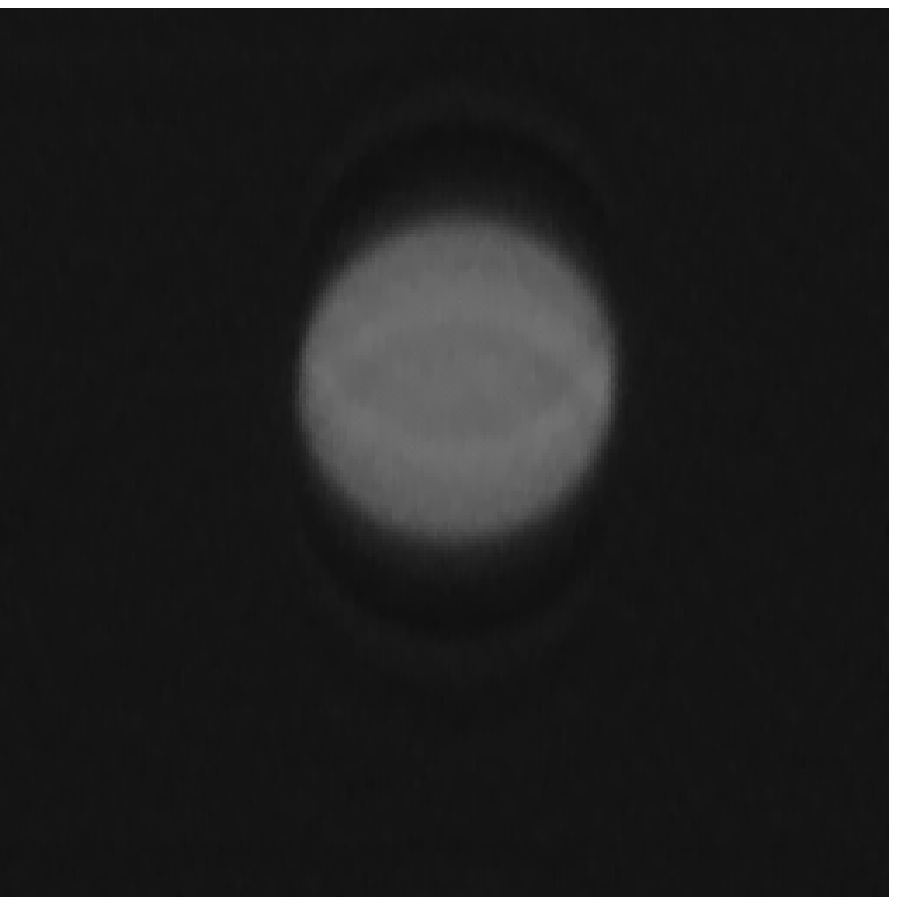}%
 \subfiglabelright{e}&
\includegraphics[width=0.24\textwidth]{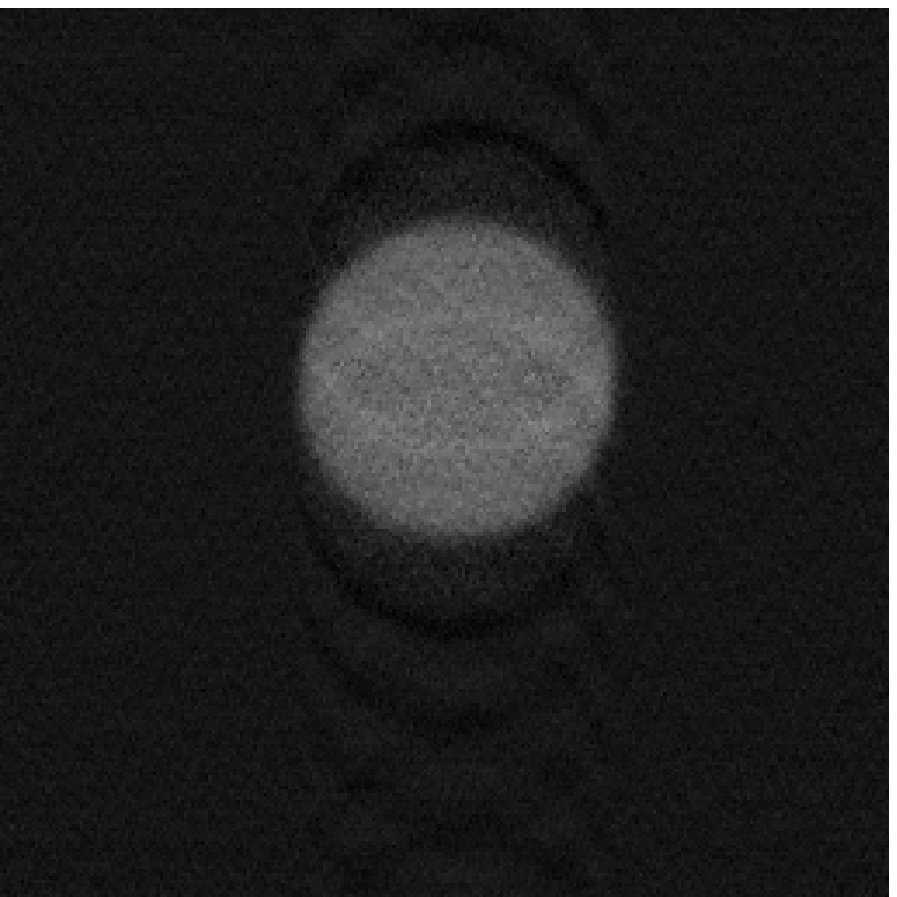}%
 \subfiglabelright{f}&
\includegraphics[width=0.24\textwidth]{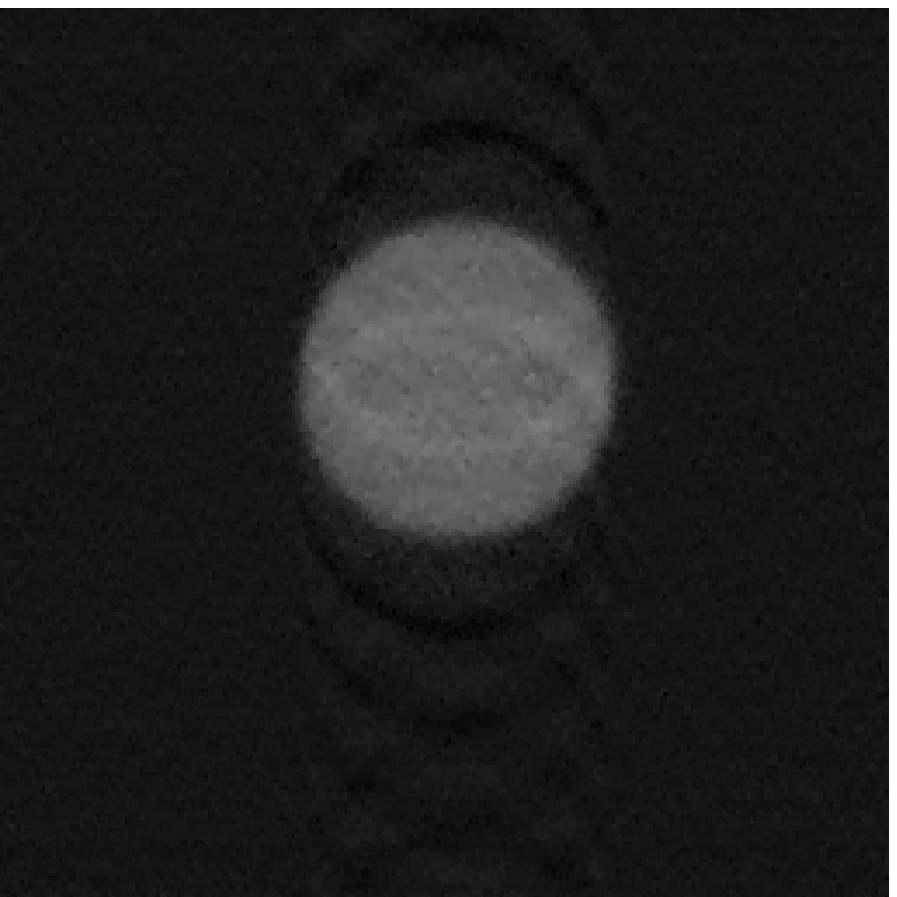}%
 \subfiglabelright{g}&
\includegraphics[width=0.24\textwidth]{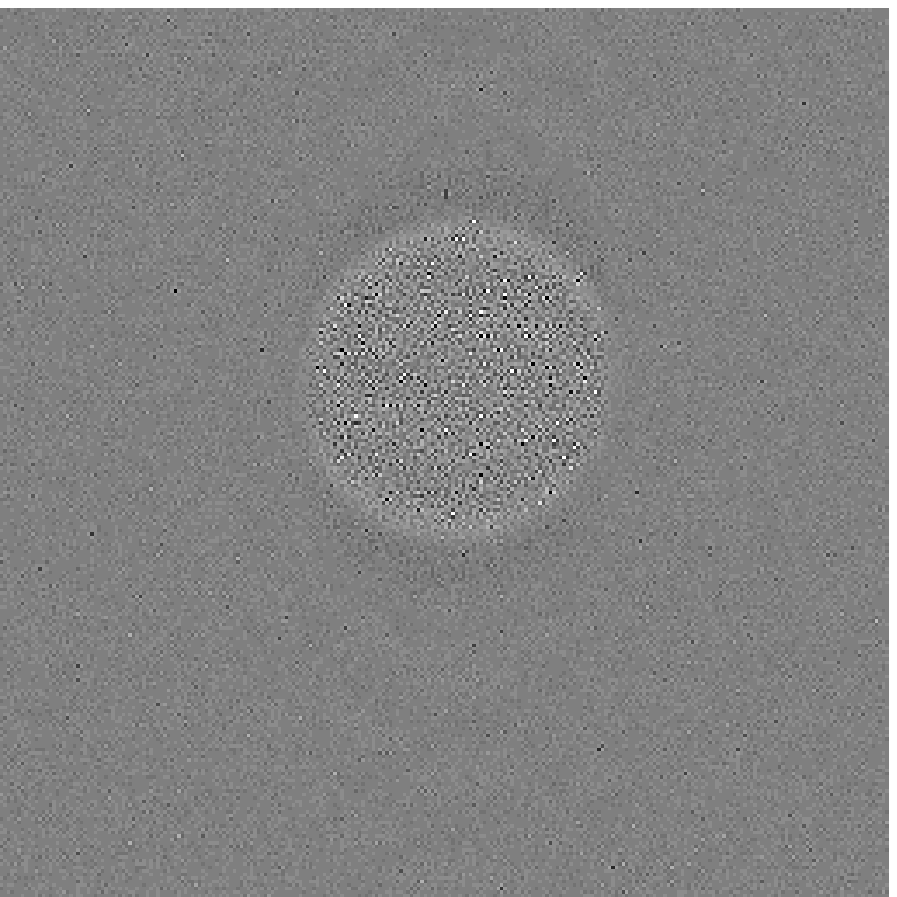}%
 \subfiglabelright{h}
\end{tabular}}
\caption{\label{fig-fidu}
\textbf{Top row, left to right:} 
\textbf{(a)}
Original image ($256\times256$ pixels) with motion blur in 
vertical direction (blur length: 27 pixels). 
Courtesy of Datacon GmbH. --
\textbf{(b)} 
Filtered by Wiener filter, $K=0.006$. --
\textbf{(c)} 
30 iterations of RL. --
\textbf{(d)}
KF method, $\lambda=100$, $\beta$ from $1$ to $128\sqrt2$ in multiplicative
steps of $2\sqrt2$, $1$ iteration per level.
\textbf{Bottom row, left to right:}
\textbf{(e)}
30 iterations of RRRL, $\alpha=0.003$. --
\textbf{(f)}
WR\textsuperscript3L with 5 iterations of RRRL. --
\textbf{(g)}
WR\textsuperscript3L with 30 iterations of RRRL. --
\textbf{(h)}
Difference image of (b) and (f), range $[-10,10]$
rescaled to $[0,255]$.}
\end{figure*}

\subsection{Restoration Quality for Real-World Images}
We turn now to real-world examples, taken under conditions similar
to the industrial production context that our development is 
directed at.
From here on, we have to rely on visual comparisons since a 
ground truth from which SNR could be computed is no longer 
available. Also, no exact knowledge on the noise distribution is 
available. On the other hand, the motion parameters determining
the blur can be adjusted in these setups, such that the
point-spread function is known. Moreover, the setting with objects
being imaged before a uniform background largely removes boundary 
artifacts -- in particular, periodic continuation is unproblematic
in this case.

\begin{figure*}[t!]
\centerline{\begin{tabular}{@{}c@{~}c@{~}c@{~}c@{~}c@{}}
\includegraphics[width=0.19\textwidth]{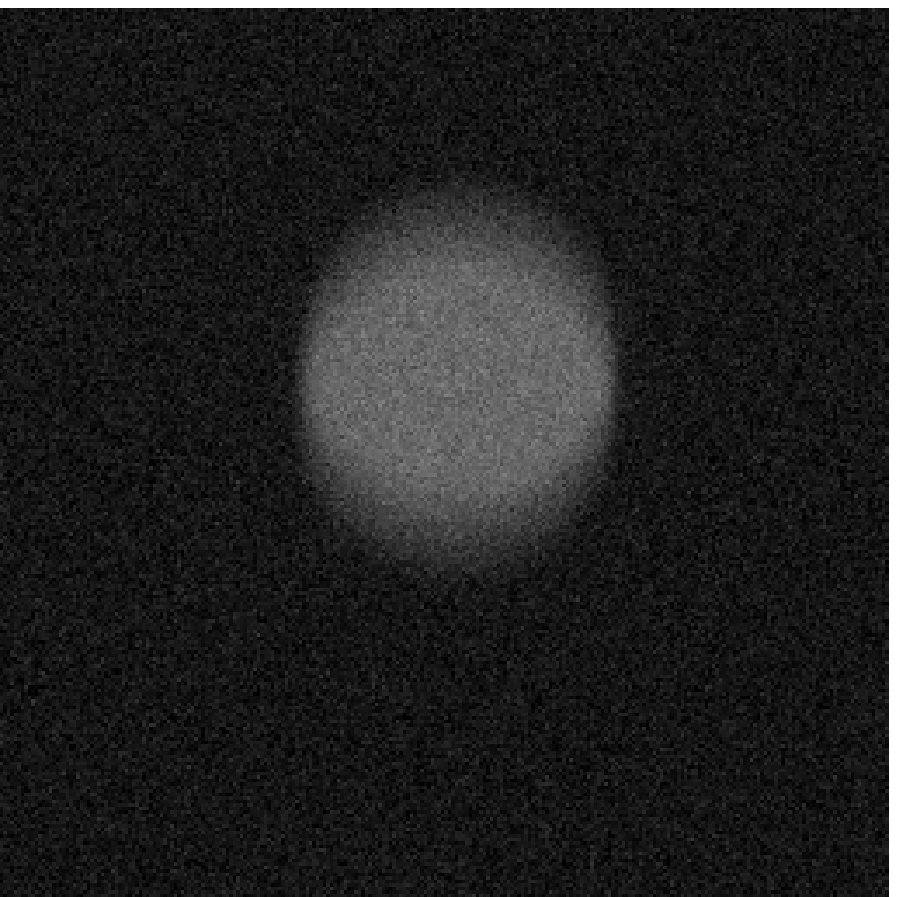}%
 \subfiglabelright{a}&
\includegraphics[width=0.19\textwidth]{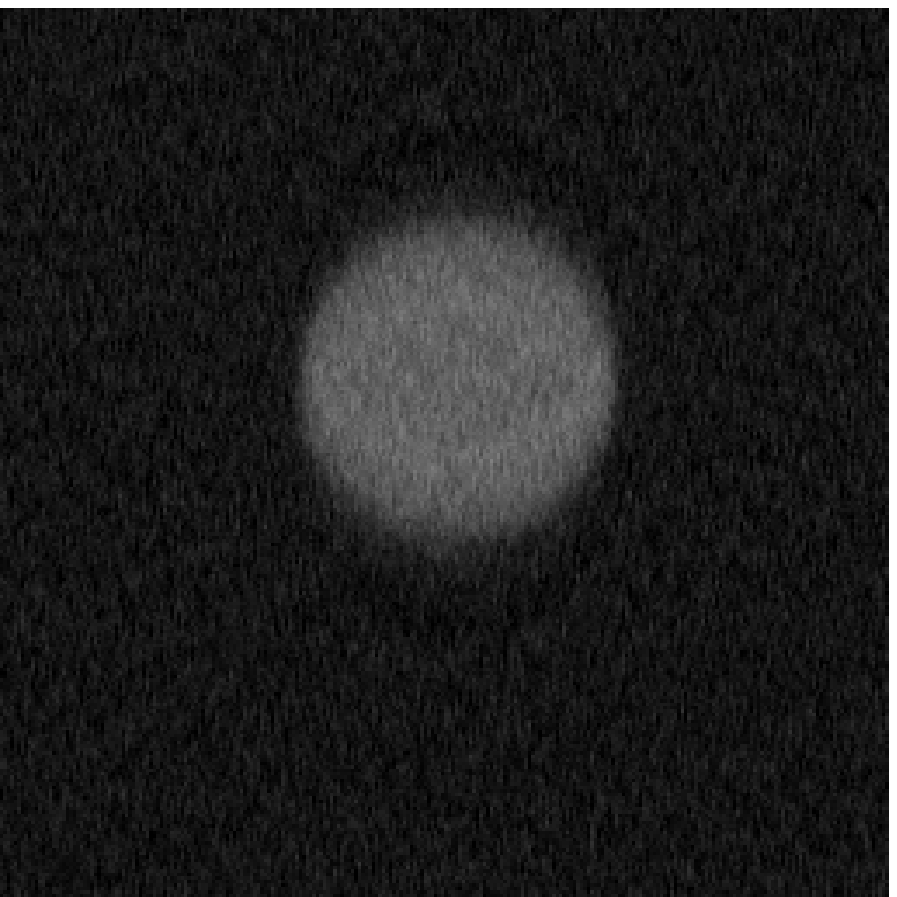}%
 \subfiglabelright{b}&
\includegraphics[width=0.19\textwidth]{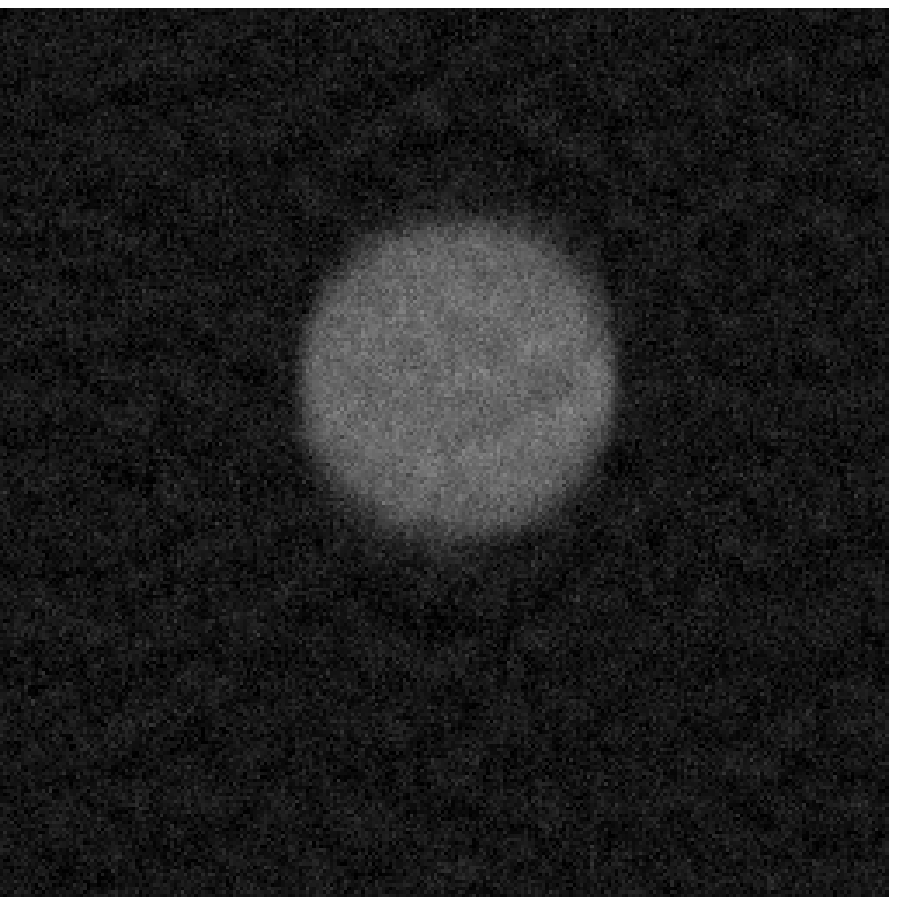}%
 \subfiglabelright{c}&
\includegraphics[width=0.19\textwidth]{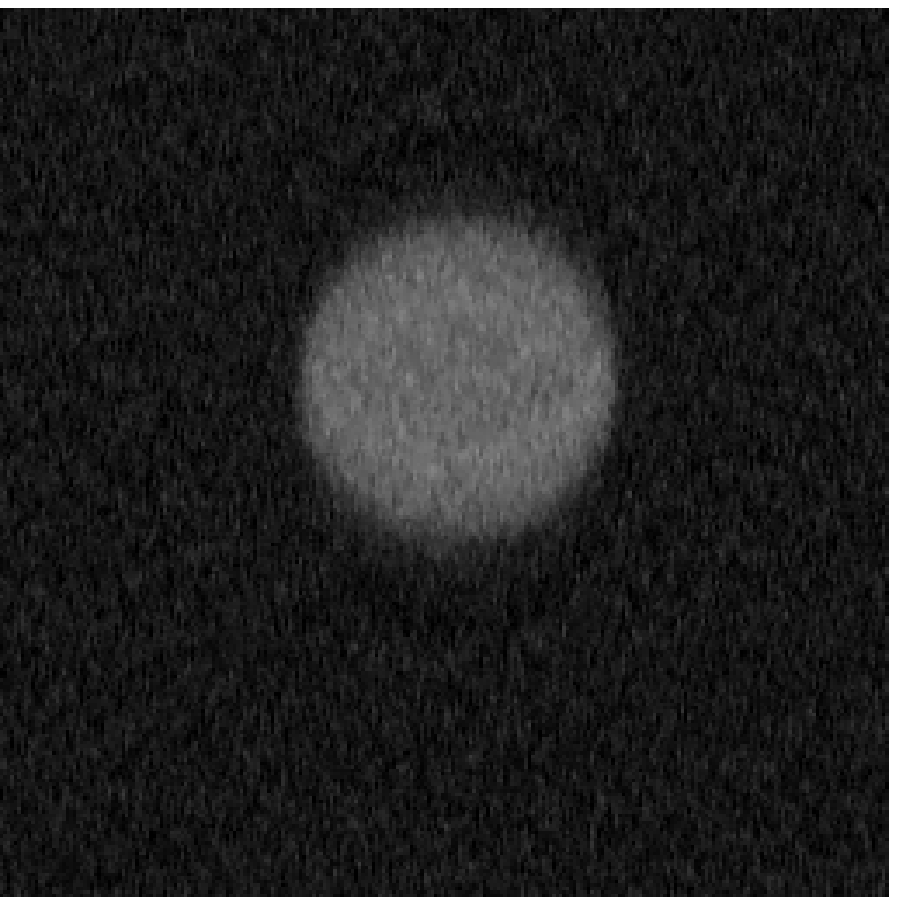}%
 \subfiglabelright{d}
\includegraphics[width=0.19\textwidth]{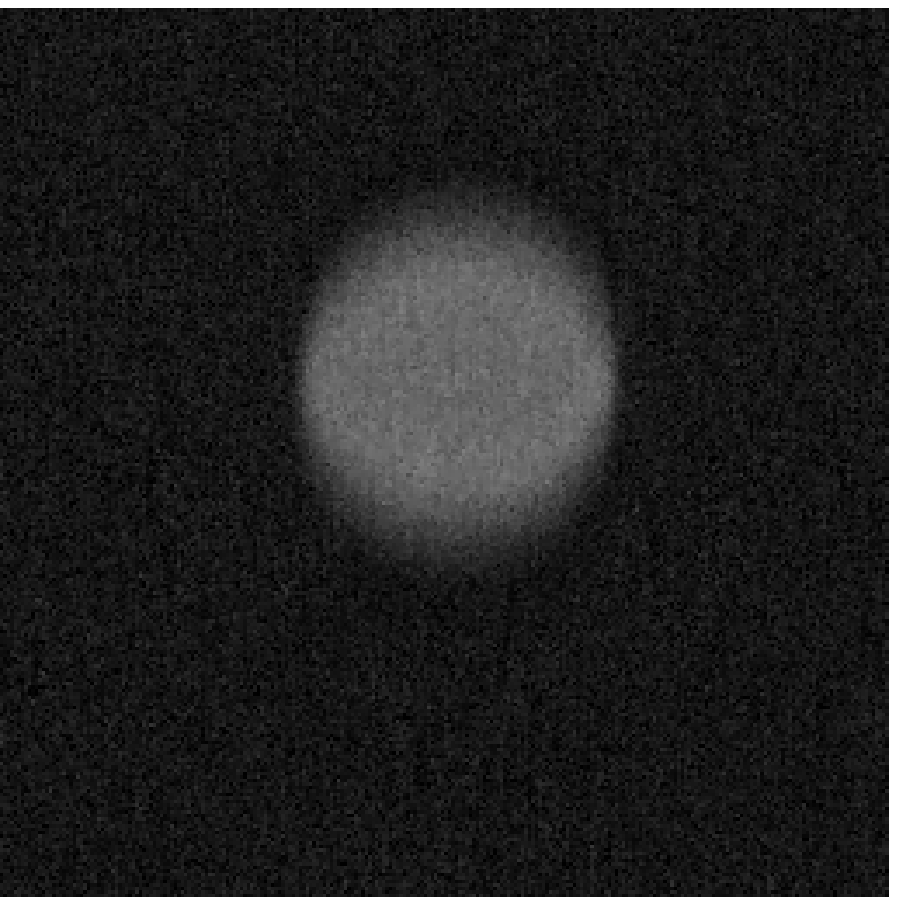}%
 \subfiglabelright{e}
\end{tabular}}
\caption{\label{fig-fidugauss}
\textbf{Left to right:} 
\textbf{(a)}
Original image from Fig.~\ref{fig-fidu}(a) with Gaussian noise,
$\sigma=10$. --
\textbf{(b)} 
Filtered by Wiener filter, $K=0.06$. --
\textbf{(c)}
Filtered by KF, $\lambda=3$, 
$\beta$ from $1$ to $128\sqrt2$ in multiplicative
steps of $2\sqrt2$, $1$ iteration per level.
\textbf{(d)} 
Wiener filter and 5 iterations of RRRL, $\alpha=0.003$. --
\textbf{(e)}
Only 5 iterations of RRRL.}
\end{figure*}

In Figs.~\ref{fig-tooth} and \ref{fig-fidu} we present two real-world 
test images with linear uniform motion blur. 
Deblurring results using 
Wiener filter, RL, KF, RRRL visually confirm the findings
from our synthetic blur experiments. The Wiener filter
allows for a visually favourable sharpening that can be equalled
by RL and RRRL only after about 30 iterations. However, 
amplified noise and artifacts appear more prominent in the Wiener 
filter result.
The combination WR\textsuperscript3L is studied in
Subfigures~(f--h) of both figures. Visually, the difference between 
the Wiener filter result, Fig.~\ref{fig-tooth}(b)/\ref{fig-fidu}(b),
and the result after additional 5 iterations of RRRL, Subfig.~(f),
appears to be small. The exaggeration with 30 iterations
in Subfig.~(g) and the difference image in Subfig.~(h)
demonstrate that there is an improvement, though:
the RRRL 
iterations keep the good initial sharpening of the Wiener filter, 
while noise and artifacts are reduced.
Ringing artifacts and noise are slightly more prominent in the KF 
result than in the RRRL or WR\textsuperscript3L results, which 
underlines that robust data terms are indeed advantageous in real-world
data.

The effect of the combination WR\textsuperscript3L becomes even
more evident in Figure~\ref{fig-fidugauss},
where the original image has been degraded by additional Gaussian
noise: Here, it is clear that the combination of Wiener filter
and RRRL combines the sharpness achieved by a single Wiener filter
step (but not by RRRL alone) with a visible noise reduction.

Summarising the observations of this section, the combination
of Wiener filter with at least five subsequent iterations of
RRRL provides a reasonable restoration quality. Our efficiency
considerations will therefore focus on this method.

\section{Efficient Implementation}
In this section, we discuss aspects of the algorithmic realisation
of the deconvolution methods under consideration.
Wiener filter, RL, RRRL and the combined WR\textsuperscript3L
were implemented in three scenarios from general 2D blur down to
1D linear motion blur. 

For WYYZ and KF only general 2D implementations were done to allow 
comparisons. 
Our strategies to derive more efficient 1D algorithms from the 
Wiener and RL filter family cannot be transferred straightforward
to the WYYZ and KF algorithms due to the different structure of their
iterations, in which the 2D regularisation is intertwined with
the 2D Fourier iteration step.

While our focus is on single-threaded CPU computation, we did also
multi-threaded CPU and GPU implementations of some variants,
which will also shortly be introduced.

\subsection{Numerics}
\label{sec-numimp}
The numerical realisation of all deconvolution methods discussed
is mostly straightforward, specifications
being necessary for the Fourier transform, convolution operations 
and the diffusion terms. 
In the case of RRRL, also the computation
of the information divergence \eqref{infdiv} deserves special
attention.{\sloppy\par}

\paragraph{Fourier transform.}
For full control over implementation details, we used our own
implementation of the Fast Fourier Transform (FFT) for power-of-two
image dimensions, based on \cite[Par.~4.4.1.3]{Bronstein-Book79},
with adaptations to the real-valuedness of image data. 
The values of the complex exponential were precomputed once for
all Fourier transforms during the program run.

\paragraph{Convolution.}
Here we considered either realisations via 
the Fourier domain, or spatial-domain convolution by direct
summation. 

\paragraph{Diffusion terms.}
These were discretised using standard central
difference approximations.

\paragraph{Function evaluations.} In the RRRL iteration,
the information divergence \eqref{infdiv} is expensive to compute 
directly due to the logarithms. For this
reason, values of $r_1(s)$ in the range $s\in[0,65]$ were
stored in a lookup table. Values of $r_f(s)$ were then calculated
by $r_f(s) = f\,r_1(s/f)$, using a linear approximation for
$r_1$ in $(65,\infty)$. (In our test examples, the latter
approximation did in fact not occur.)

In the KF(-S) algorithm the solution of a polynomial equation
can be replaced by a lookup table. In \cite{Krishnan-nips09},
speedups from about $1.7$ (for $256\times256$ images) to $4$ 
(for large images) were obtained in this way. In our implementation,
we stick to the slower analytic solution. However, by comparing
with the WYYZ algorithm the possible speedup can be estimated.

\subsection{Boundary Treatment}
Since blur operations involve considerable transport of
information across the image boundary, artifacts
near the image boundary are an issue in deconvolution. 
These artifacts are especially strong when simple boundary
conditions like zero-padding, constant or periodic continuation
introduce massive model violations. Although advanced boundary 
treatment schemes are available that allow to reduce these 
artifacts considerably, see e.g.\ \cite{Chan-IJIST05},
these are computationally expensive. 

For performance reasons, we decide to use constant continuation
for spatial convolution operations, and the natural periodic
continuation for Fourier-based operations, and tolerate the 
resulting artifacts. Note that in the application setting of
Figs.~\ref{fig-tooth}--\ref{fig-fidugauss} where
objects are photographed in front of uniform backgrounds,
almost no artifacts are introduced anyway.

\subsection{Algorithmic Optimisations}
Concluding from the qualitative tests, it is desirable to 
perform Wiener filtering plus at least five iterations of
RRRL for a practically useful deconvolution in the real-time
environment. Concerning the image size, we restrict ourselves
to powers of two in order to profit most from the Fast Fourier
Transform (FFT). Depending on the exact algorithmic variant this 
restriction can be eased in application.

In our examples, an image size of $256\times256$ is appropriate 
to include a suitable region of interest for object detection or 
localisation. For more limited applicability, also images of size 
$128\times128$ may be considered.

We discuss now algorithmic optimisations that can be used to 
achieve the desired deconvolution in the so defined setting.
We will distinguish herein between uniform linear motion and
non-uniform linear motion, and compare both to a general 2D
space-invariant blur situation.

\paragraph{General 2D setting.} It is clear that in the general
2D blur scenario, Wiener filtering requires a two-di\-men\-sio\-nal
Fourier transform and inverse transform, which are
implemented by FFT.

In the iterative algorithms (RL, RRRL), convolutions can either be 
computed by direct numerical integration in the spatial domain,
or again via the Fourier domain. In the latter case, once more
a 2D FFT is necessary. As the complexity of the spatial
domain convolution is linear in the image size and PSF size,
while the FFT implementation is log-linear in the image size
only, spatial domain convolution may
be superior to FFT for very small PSF but FFT will
dominate for large convolution kernels.

\paragraph{Non-uniform linear motion.} In this case the 
blur is characterised by a point-spread function with 1D support.
Assuming that it is aligned with a coordinate axis of the imaging
device (say the vertical one), the Wiener
filter needs to applied only in columns. Thus, a 1D FFT is
sufficient, saving up to half the numerical cost of the Fourier
transform.

Equally, only 1D convolutions are needed in the iterative
method. If these are carried out in 1D only, the program logic
is slightly simplified compared to the 2D case but the number of
multiplications and additions in evaluating the integral will
be comparable to that of a 2D implementation for equal PSF
size. In contrast, a Fourier-based realisation will again
profit from using 1D instead of 2D FFT.

\paragraph{Uniform linear motion.}
The point-spread function in this case still is 1D, but it
takes the special form of a box filter, i.e.\ it is
constant throughout its support. (This applies exactly when
the blur length is integer. In the case of a non-integer
blur length, one has to allow for single pixels of
smaller weight at one or both ends of the kernel.)

For the Wiener filter, this setting does not offer any specific 
advantage over the general 1D case.
The same holds true for the convolutions in the iterative
method when computed via the Fourier domain.

However, the spatial domain convolution can be made dramatically
more efficient in this case by an efficient box filtering
algorithm \cite{McDonnell-CGIP81}. 
In the case of a vertical linear blur this algorithm works
as follows: In each column, the convolution of the first pixel
is computed by direct summation (complexity linear w.r.t.\
the kernel size). Then the sliding window is shifted one pixel
at a time, such that one pixel enters the window while one
pixel leaves it. So the sum inside the window is updated in
constant time by adding one grey-value and subtracting another one.
If non-integer blur length is admitted, the update step involves 
at most two additions and two subtractions.
The overall complexity of the update part is therefore linear
in the image size, and the total complexity of the algorithm with
blur kernel size $m$ on an image of $n_x$ columns and $n_y$
rows amounts to $\mathcal{O}(n_x\cdot(n_y+m))$.

\subsection{Implementation and Technical Optimisation}
All CPU programs were written in C, 
and compiled using gcc 4.6 with optimisation level O2. 
With O3 some run times were further reduced by a few percent
while others even increased slightly. Also more specific compiler
optimisation settings did not significantly improve performance.
However, efficiency was improved by standard source 
code optimisation techniques like inlining, loop merging, blocking
(for cache optimisation).
Wherever possible, array access operations were avoided using
auxiliary variables. One-dimensional arrays were used to store
image data.

\subsection{Parallelisation}
While our focus is on single-threaded CPU computation,
we consider two exemplary parallel implementations to
assess the possible gains by parallelisation on multicore
CPUs and GPUs.

\paragraph{Multi-threaded CPU implementation, 1D.}
We imple\-men\-ted the version of RRRL for general 1D motion
blur for multi-threaded computation on a multi-core CPU
using the standard \texttt{pthreads} library.
The Wiener filter was not parallelised in this setting because
it contributes little to the overall run-time.
{\sloppy\par}

The sharpening term of the RRRL iteration nicely decomposes
in this case in columns parallel to the point-spread function,
and is easily parallelised in this way.
In contrast, the smoothing term is still made up by 2D differential
expressions, so the columns interact with each other in the 
derivative computation.
Since the overall computational cost of the sharpening
term equals several times that of the smoothing term, an
almost balanced workload is achieved on CPUs with 4\ldots8 cores 
if the regulariser is computed in the parent thread, while
the sharpening term is distributed to the remaining cores in
parallel threads. 

To reduce the overhead of creating and terminating threads,
the parallel worker threads for the sharpening term
are started before the first RRRL iteration, and not terminated
before all iterations are done. Mutexes are used to synchronise
the updates between worker threads and parent thread in each
iteration.

\paragraph{GPU implementation, 2D Fourier.}
We also im\-ple\-men\-ted the WR\textsuperscript3L algorithm for
general 2D point-spread functions using the CUDA 4.0 framework
for Nvidia GPUs. Both the Wiener filter and the convolutions
in the RRRL iteration are performed using CUDA's built-in
Fast Fou\-rier transforms. Efficiency of parallel access to
some data (namely, the point-spread function) was improved by
using texture memory.

\section{Experiments}
We measure the performance of deconvolution algorithms suitable
for the three scenarios described in the previous section.
For this, we used the \texttt{gettimeofday()} function since it
states real-world run times within the context of the running
system instead of pure process time, and easily allows measurements
for any particular portion of an algorithm. The downside is that
due to other activities (system processes etc.) run times will 
display considerable variation -- we used a standard Linux system
without any specific real-time scheduler.
It is therefore necessary to
consider statistics over many program runs.
We report
therefore averages, standard deviations and extremes from 100 
subsequent program runs.

Not included in our measurements is the time for loading and storing
images. The rationale for this is that in time-critical industrial 
applications, image data would anyway be transferred into the memory 
directly from the imaging device.

Also not included is the time for precomputing auxiliary data for the
Fourier transform. This is based on the assumption that this can be 
done once for a large number of equally sized images to be processed 
in an application context.

\subsection{Single-Threaded Wiener+RRRL}
\paragraph{Comparison across 1D/2D blur scenarios.}
We choose as our test case the image from Fig.~\ref{fig-fidu}
($256\times256$ pixels)
with the uniform linear motion blur kernel of length $27$,
for which all algorithms could equally be applied.
We remark that for the algorithms chosen (box filter, Fourier
convolution/Wiener filtering) the computational cost does not
or only slightly depend on the size of the blur kernel.
In all cases, we compute Wiener filtering plus five iterations
of RRRL.

\begin{table*}[b!]
\caption{\label{tab-rt256}
Run times for sharpening the image from Fig.~\ref{fig-fidu}
($256\times256$)
with Wiener filter followed by 5 iterations of RRRL. 
Details see text. 
Boldface figures: average times $\pm$ estimated standard deviation,
figures in brackets: minimum/maximum. Statistics from 100 program
runs for each method.}
\centerline{%
\begin{tabular}{|l||c|c|c||c|}
\hline
Algorithm&\multicolumn{4}{|c|}{Run time (ms)}\\
&Wiener&RRRL iteration&RRRL total&Total\\
\hline
1D with box filter&%
\textbf{3.8}${}\pm{}$2.3&%
\textbf{4.8}${}\pm{}$0.3&%
\textbf{24.3}${}\pm{}$0.5&%
\textbf{28.1}${}\pm{}$2.4\\
&{(2.8\dots12.2)}&%
{(4.3\dots6.2)}&%
{(22.8\dots26.8)}&%
{(25.8\dots37.1)}\\
\hline
1D with Fourier&%
\textbf{3.1}${}\pm{}$0.7&%
\textbf{10.9}${}\pm{}$0.3&%
\textbf{54.6}${}\pm{}$1.1&%
\textbf{57.7}${}\pm{}$1.5\\
&(2.8\dots8.3)&%
(10.2\dots11.9)&%
(52.6\dots57.0)&%
(55.6\dots64.9)\\
\hline
2D with Fourier&%
\textbf{14.1}${}\pm{}$4.1&%
\textbf{20.5}${}\pm{}$0.3&%
\textbf{102.7}${}\pm{}$0.5&%
\textbf{116.7}${}\pm{}$4.1\\
&(11.7\dots27.3)&%
(20.0\dots21.4)&%
(101.5\dots103.8)&%
(113.2\dots130.0)\\
\hline
\end{tabular}}
\end{table*}

\begin{table*}[b!]
\caption{\label{tab-rt512}
Run times for sharpening images of different sizes
with 1D Wiener filter followed by 5 iterations of RRRL with box
filter algorithm. 
Details see text. 
Boldface figures: average times $\pm$ estimated standard deviation,
figures in brackets: minimum/maximum. Statistics from 100 program
runs for each image size.}
\centerline{%
\begin{tabular}{|l||c|c|c||c|}
\hline
Image size&\multicolumn{4}{|c|}{Run time (ms)}\\
&Wiener&RRRL iteration&RRRL total&Total\\
\hline
$128\times128$&%
\textbf{1.5}${}\pm{}$1.1&%
\textbf{1.4}${}\pm{}$1.1&%
\textbf{7.3}${}\pm{}$3.4&%
\textbf{8.7}${}\pm{}$4.4\\
&{(0.6\dots3.4)}&%
{(0.9\dots4.7)}&%
{(4.6\dots15.8)}&%
{(5.2\dots18.7)}\\
\hline
$256\times256$&%
\textbf{3.8}${}\pm{}$2.3&%
\textbf{4.8}${}\pm{}$0.3&%
\textbf{24.3}${}\pm{}$0.5&%
\textbf{28.1}${}\pm{}$2.4\\
&{(2.8\dots12.2)}&%
{(4.3\dots6.2)}&%
{(22.8\dots26.8)}&%
{(25.8\dots37.1)}\\
\hline
$512\times512$&%
\textbf{14.2}${}\pm{}$0.2&%
\textbf{24.5}${}\pm{}$0.9&%
\textbf{123.6}${}\pm{}$0.8&%
\textbf{137.8}${}\pm{}$0.8\\
&{(14.0\dots15.8)}&%
{(23.6\dots27.1)}&%
{(121.6\dots126.4)}&%
{(135.7\dots140.4)}\\
\hline
\end{tabular}}
\end{table*}

\begin{table*}[b!]
\caption{\label{tab-rt-kfwang}
Run times for sharpening the image from Fig.~\ref{fig-fidu}
($256\times256$)
with different methods suitable for general 2D blurs. 
Details see text. 
Boldface figures: average times $\pm$ estimated standard deviation,
figures in brackets: minimum/maximum. Statistics from 100 program
runs for each method.}
\centerline{%
\begin{tabular}{|l||c|}
\hline
Algorithm&Run time (ms)\\
\hline
KF with analytic solver, 6 iterations&
\textbf{372.2}${}\pm{}$3.2\\
(parameter regime from \cite{Krishnan-nips09})&(366.1\dots384.4)\\
\hline
WYYZ, 6 iterations&
\textbf{82.0}${}\pm{}$2.1\\
(parameter regime adapted to \cite{Krishnan-nips09})&(73.2\dots88.8)\\
\hline
WR\textsuperscript3L, 5 iterations&
\textbf{116.7}${}\pm{}$4.1\\
(2D Fourier)&(113.2\dots130.0)\\
\hline
\end{tabular}}
\end{table*}

\begin{table*}[b!]
\caption{\label{tab-rt-mt-gpu}
Run times for sharpening the image from Fig.~\ref{fig-fidu}
($256\times256$)
with Wiener filter followed by 5 iterations of RRRL in
exemplary parallel implementations.
Details see text. 
Boldface figures: average times $\pm$ estimated standard deviation,
figures in brackets: minimum/maximum. Statistics from 100 program
runs for each method.}
\centerline{%
\begin{tabular}{|l||c|c|c||c|}
\hline
Implementation&\multicolumn{4}{|c|}{Run time (ms)}\\
&Wiener&RRRL iteration&RRRL total&Total\\
\hline
Multi-threaded CPU&%
\textbf{4.0}${}\pm{}$1.7&%
\textbf{5.4}${}\pm{}$2.4&%
\textbf{27.8}${}\pm{}$5.6&%
\textbf{31.8}${}\pm{}$6.4\\
(1D Fourier)&(2.9\dots13.4)&%
(2.3\dots17.3)&%
(17.9\dots42.7)&%
(20.9\dots48.2)\\
\hline
GPU (CUDA)&%
\textbf{0.35}${}\pm{<}$0.01&%
\textbf{1.36}${}\pm{}$0.01&%
\textbf{6.80}${}\pm{}$0.01&%
\textbf{7.15}${}\pm{}$0.01\\
(2D Fourier)&(0.34\dots0.36)&%
(1.34\dots1.39)&%
(6.78\dots6.84)&%
(7.13\dots7.19)\\
\hline
\end{tabular}}
\end{table*}

Run times were measured for single-threaded computation (thus
using one core) on an AMD Phenom II X6 1100T running at 3.3 GHz.
Statistics for the Wiener step, single RRRL iterations, 
total run time of the five RRRL iterations and overall time are 
given in Table~\ref{tab-rt256}. 
Note that the Fourier transform of the point-spread function
is computed once in the Wiener filter step and re-used if
necessary in the RRRL iterations.

From this table it is evident that the proposed filtering
procedure can be carried out reliably in less than 50 ms on
$256\times256$ grey-value images with uniform linear motion
blur. For the more general blur scenarios, the computational
expense still prevents real-time performance in the
single-threaded setting, although even here favourable run-times 
are achieved.

\paragraph{Different image sizes.}
In Table~\ref{tab-rt512} run times for the fas\-test algorithm
(1D Wiener, RRRL with box filter) are shown for three different
image sizes, all with the same blur kernel size.
As can be seen, the measured run times for the RRRL iteration
scale by a factor $3\dots5$ for each quadrupling of the image 
size, reflecting the essentially linear complexity. In contrast,
averaged run times for the Wiener filter step (actually log-linear)
multiply only by $2.5\dots3.5$ for each scale step.

This effect can be traced back to outliers with large execution
times that occur in the Wiener filter step of some program runs 
and sometimes also in the first RRRL iterations. These outliers 
seem to be caused by cache effects and amount to an almost 
constant-time overhead on the average times, and thus the misleading 
impression of a sub-linear scaling of the Wiener filter step.
Indeed, when for testing the Wiener filtering step is performed 
twice, the second step displays less outliers and thus a lower
average.
Similarly, the first RRRL iteration in each program run is
slightly slower than the subsequent iterations (about 0.5 ms
for $256\times256$, about 2 ms for $512\times512$).
The influence of the outliers is particularly pronounced for
$128\times128$ images, as indicated by the large standard
deviations. It is evident that measurements for these short
execution times are less reliable than for the larger images.

\subsection{Comparison to WYYZ and KF}

We test the KF and WYYZ algorithms on the same $256\times256$
test case as the WR\textsuperscript3L algorithms. 
Results are shown in Table~\ref{tab-rt-kfwang}.
As these algorithms are implemented in the full 2D setting, 
the proper WR\textsuperscript3L value for comparison comes from
the last row of Table~\ref{tab-rt256}.

The KF algorithm ($\alpha=2/3$) with analytic solver is tested
in the $\beta$ parameter regime recommended in 
\cite{Krishnan-nips09} ($\beta$ scaling in powers of $\sqrt8$ 
from $1$ to $256$, one inner iteration per level) which totals to
6 iterations. The resulting run-time is in good agreement with
the original value of 0.7\,s from \cite{Krishnan-nips09}
when compensating for the different CPU clocks and image dimensions.

For the WYYZ algorithm, we use the same parameter regime.
Since WYYZ differs from KF just by having a cheap shrinkage
step where the expensive polynomial equation is solved in KF,
the so measured run times give a reliable lower bound to
the run time that KF could reach with the lookup table instead
of the polynomial solver.
Note, however, that \cite{Wang-SIIMS08} suggests finer scaling 
steps for $\beta$ and substantially more iterations per level.

It can be seen from these figures that WR\textsuperscript3L,
albeit not the fastest algorithm, is competitive in terms of
speed.

\subsection{Parallel Implementations of Wiener+RRRL}

\paragraph{Multicore CPU computation.}
The gap between the run-times displayed in Table~\ref{tab-rt256}
for the general 1D setting for images of size $256\times256$
and our goal of real-time computation is not very large. Therefore
it is interesting to see whether this gap could be bridged by
multithreaded computation on a recent multicore CPU.

We tested this on the Phenom X6 hexacore machine described in
the previous subsection. In our parallelised implementation
of the RRRL component the regularisation was computed in the
main thread, and the sharpening terms distributed to five threads
such that all six cores could be used. Run-time results are shown
in the first row of Table~\ref{tab-rt-mt-gpu}. In our test setting
with just five iterations, the effective speed-up factor for the
RRRL computation is only about $2$. 

A closer look at the run-times
of single iterations reveals that typically the first two to three
iterations are slower than the following ones.
In fact, the averages and standard deviations for the five iterations
are $(7.0\pm3.5)\,\mathrm{ms}$, $(7.3\pm1.8)\,\mathrm{ms}$,
$(4.7\pm1.3)\,\mathrm{ms}$, $(4.2\pm0.9)\,\mathrm{ms}$, and 
$(3.9\pm1.0)\,\mathrm{ms}$, respectively.
It appears that the overhead caused by creating and managing threads 
chips away a lot of the possible gain in performance for such short
computational tasks.
Indeed, when running much more iterations, the average computation time
per RRRL iteration stabilises in the range of $(3\ldots4)\,\mathrm{ms}$.
The workload is well balanced between cores, as indicated by the fact that
\texttt{top} consistently shows CPU loads of about $580\,\%$ for the
process.

In spite of the modest speed-up factor, the multicore computation
achieves the goal of performing WR\textsuperscript3L with 5 iterations
on $256\times256$ pixels within $50\,\mathrm{ms}$.

\paragraph{GPU computation.} We tested our GPU implementation
of WR\textsuperscript3L for general 2D point-spread functions on an nVidia
GT-440 graphics card featuring 96 cores at a clock rate of 
$1620\,\mathrm{MHz}$.
The net computation times of the Wiener filter and RRRL iterations, and the
entire WR\textsuperscript3L are shown in the second row of 
Table~\ref{tab-rt-mt-gpu}.

Firstly, standard deviations of these figures are much lower than for
the CPU computations. This can be attributed to the fact that there are 
almost no other processes in the system that exploit the computing capabilities
of the graphics card, and could thus interfere with the computation.

Secondly, the time measurements show that GPU computation enables the general 
2D deconvolution of $256\times256$-pixel images to be performed reliably under 
our real-time constraint ($50\,\mathrm{ms}$); even the computation
for $512\times512$-pixel images appears feasible. 

Real-time-capable GPU deconvolution for general 2D blurs has already
been devised in \cite{Klosowski-spie11}, compare also the non-blind 
deconvolution part in the framework of \cite{Hirsch-iccv11}. Both papers
use the KF algorithm \cite{Krishnan-nips09} with the same parameter regime
that we have used in Section~\ref{sec-qual}. We juxtapose therefore
our speed measurements with those reported in \cite{Klosowski-spie11}.

The average total run-time from Table~\ref{tab-rt-mt-gpu} amounts to
about 9.16 megapixels per second of single-channel computation, i.e.\
about 58.9 pixels per core per $10^6$ clock cycles.
Since our implementation is not applicable to arbitrary image sizes,
the test case from \cite{Klosowski-spie11} is the one where the image
size optimally fits their algorithm. For this scenario, 
\cite{Klosowski-spie11} reports about 13 megapixels per 
second (single-channel). The value refers to an nVidia GTX 260 graphics card 
with 192 cores (according to nVidia's official specification; 
\cite{Klosowski-spie11} states 216) and a $1242\,\mathrm{MHz}$ clock, 
which means that about 54.5 pixels are processed per core per $10^6$ clock 
cycles. Given the neglection of other influence factors, this
can only serve as a rough comparison, but it demonstrates that the 
computational efficiency of both approaches is in the same range.

With its slight advantage in restoration quality in a number of settings
demonstrated in Section~\ref{sec-qual}, WR\textsuperscript3L lends itself
therefore as an attractive candidate also for general 2D deconvolution
under real-time conditions on the GPU.

\medskip
Nevertheless, some words of care must be said. In tune with our decision to
exclude loading and storing images from the time measurements, the transfer
between main memory and graphics card memory is not contained in the mentioned
run-time figures. However, including these data transfers does not 
substantially change the picture. In our example, these transfers total to 
about $0.4\,\mathrm{ms}$.

Let us also revisit our other exclusion, precomputation of auxiliary data 
for Fourier transforms. In our CPU implementation, inclusion of these 
computations would make the computations more expensive, but not 
dramatically so.
In the GPU setting with CUDA's built-in Fourier transform, the precomputation
of Fourier transform plans is fairly expensive. We measured run-times of about
$120\,\mathrm{ms}$ for this step; however, it is open to some question whether 
this time is exaggerated by including some initialisation overhead. 
Of course, for the price of the expensive Fourier plan construction we get a
degree of generality that is not with our specialised CPU implementation. Very 
likely, using a more flexible off-the-shelf Fourier transform package on the 
CPU would lead to a similar shift in balance between precomputation and actual 
image processing. At any rate, for the practical efficiency of the proposed 
GPU-based deconvolution it is crucial to ensure that precomputed data are 
retained and used for multiple images.

\section{Summary and Outlook}

We have demonstrated the design of an efficient
and robust deconvolution algorithm for known space-invariant blur. 
By combining Wiener filtering as a first step with a small
number of iterations of robust and regularised Richardson-Lucy
deconvolution \cite{Welk-tr10} a reasonable deconvolution quality
is achieved at a fairly low computational expense.
We improved this basic method by algorithmic optimisations for 
specific blur scenarios, in particular fast box filtering 
\cite{McDonnell-CGIP81} for uniform linear motion blur.
In this case, real-time performance was reached for moderate image 
sizes in single-threaded CPU computation. To our knowledge, there 
has been no comparable framework so far for CPU-based real-time 
image deconvolution, even if it is only in a specific setting.

Exemplary implementations demonstrated also
that comparable real-time performance can also be achieved for 
general 1D blur by multi-threaded computation on a contemporary multi-core CPU,
and for 2D blur using GPU computation.

Ongoing work is directed at further specific blur settings
as well as a more systematic investigation of efficient parallel 
implementations.
Also, the comparison with existing real-time-capable GPU-based deconvolution 
in terms of restoration quality and robustness deserves further consideration. 
We expect that by these efforts
the applicability of deconvolution in automatisation, quality 
inspection and further application fields will be significantly
improved. 

\section*{Acknowledgements}

It is gratefully acknowledged that work on this project was
funded by Standortagentur Tirol, Innsbruck, and done in cooperation 
with Datacon GmbH, Radfeld, and WESTCAM Projektmanagement GmbH, Mils.

\bibliographystyle{spmpsci}
\bibliography{dcvrefs}
\end{document}